\documentclass{article}
\usepackage[disable]{todonotes} 

\usepackage[utf8]{inputenc} 
\DeclareUnicodeCharacter{2032}{\ensuremath{'}}
\usepackage[T1]{fontenc}    

\usepackage{natbib}
\usepackage{environ}
\newcommand{\acksection}{\section*{Acknowledgments and Disclosure of Funding}}
\NewEnviron{ack}{\acksection\BODY}

\usepackage[verbose=true,letterpaper]{geometry}
\AtBeginDocument{
  \newgeometry{
    textheight=9in,
    textwidth=5.5in,
    top=1in,
    headheight=12pt,
    headsep=25pt,
    footskip=30pt
  }
}

\makeatletter
\renewcommand{\normalsize}{%
  \@setfontsize\normalsize\@xpt\@xipt
  \abovedisplayskip      7\p@ \@plus 2\p@ \@minus 5\p@
  \abovedisplayshortskip \z@ \@plus 3\p@
  \belowdisplayskip      \abovedisplayskip
  \belowdisplayshortskip 4\p@ \@plus 3\p@ \@minus 3\p@
}
\normalsize
\renewcommand{\small}{%
  \@setfontsize\small\@ixpt\@xpt
  \abovedisplayskip      6\p@ \@plus 1.5\p@ \@minus 4\p@
  \abovedisplayshortskip \z@  \@plus 2\p@
  \belowdisplayskip      \abovedisplayskip
  \belowdisplayshortskip 3\p@ \@plus 2\p@   \@minus 2\p@
}

\renewcommand{\large}{\@setfontsize\large\@xiipt{14}}
\renewcommand{\Large}{\@setfontsize\Large\@xivpt{16}}
\renewcommand{\LARGE}{\@setfontsize\LARGE\@xviipt{20}}
\renewcommand{\huge}{\@setfontsize\huge\@xxpt{23}}
\renewcommand{\Huge}{\@setfontsize\Huge\@xxvpt{28}}

\renewcommand{\section}{%
  \@startsection{section}{1}{\z@}%
                {-2.0ex \@plus -0.5ex \@minus -0.2ex}%
                { 1.5ex \@plus  0.3ex \@minus  0.2ex}%
                {\large\bf\raggedright}%
}
\renewcommand{\subsection}{%
  \@startsection{subsection}{2}{\z@}%
                {-1.8ex \@plus -0.5ex \@minus -0.2ex}%
                { 0.8ex \@plus  0.2ex}%
                {\normalsize\bf\raggedright}%
}
\renewcommand{\subsubsection}{%
  \@startsection{subsubsection}{3}{\z@}%
                {-1.5ex \@plus -0.5ex \@minus -0.2ex}%
                { 0.5ex \@plus  0.2ex}%
                {\normalsize\bf\raggedright}%
}
\renewcommand{\paragraph}{%
  \@startsection{paragraph}{4}{\z@}%
                {1.5ex \@plus 0.5ex \@minus 0.2ex}%
                {-1em}%
                {\normalsize\bf}%
}
\renewcommand{\subparagraph}{%
  \@startsection{subparagraph}{5}{\z@}%
                {1.5ex \@plus 0.5ex \@minus 0.2ex}%
                {-1em}%
                {\normalsize\bf}%
}

\newlength{\@neuripsabovecaptionskip}
\newlength{\@neuripsbelowcaptionskip}
\renewcommand{\footnoterule}{\kern-3\p@ \hrule width 12pc \kern 2.6\p@}
\def\@listi  {\leftmargin\leftmargini}
\def\@listii {\leftmargin\leftmarginii
              \labelwidth\leftmarginii
              \advance\labelwidth-\labelsep
              \topsep  2\p@ \@plus 1\p@    \@minus 0.5\p@
              \parsep  1\p@ \@plus 0.5\p@ \@minus 0.5\p@
              \itemsep \parsep}
\def\@listiii{\leftmargin\leftmarginiii
              \labelwidth\leftmarginiii
              \advance\labelwidth-\labelsep
              \topsep    1\p@ \@plus 0.5\p@ \@minus 0.5\p@
              \parsep    \z@
              \partopsep 0.5\p@ \@plus 0\p@ \@minus 0.5\p@
              \itemsep \topsep}

\newcommand{\@toptitlebar}{%
  \hrule height 4\p@
  \vskip 0.25in
  \vskip -\parskip%
}
\newcommand{\@bottomtitlebar}{%
  \vskip 0.29in
  \vskip -\parskip
  \hrule height 1\p@
  \vskip 0.09in%
}
\renewcommand{\maketitle}{%
  \par
  \begingroup
    \renewcommand{\thefootnote}{\fnsymbol{footnote}}
    \renewcommand{\@makefnmark}{\hbox to \z@{$^{\@thefnmark}$\hss}}
    \long\def\@makefntext##1{%
      \parindent 1em\noindent
      \hbox to 1.8em{\hss $\m@th ^{\@thefnmark}$}##1
    }
    \thispagestyle{empty}
    \@maketitle
    \@thanks
  \endgroup
  \let\maketitle\relax
  \let\thanks\relax
}
\renewcommand{\@maketitle}{%
  \vbox{%
    \hsize\textwidth
    \linewidth\hsize
    \vskip 0.1in
    \@toptitlebar
    \centering
    {\LARGE\bf \@title\par}
    \@bottomtitlebar
    \def\And{%
      \end{tabular}\hfil\linebreak[0]\hfil%
      \begin{tabular}[t]{c}\bf\rule{\z@}{24\p@}\ignorespaces%
    }
    \def\AND{%
      \end{tabular}\hfil\linebreak[4]\hfil%
      \begin{tabular}[t]{c}\bf\rule{\z@}{24\p@}\ignorespaces%
    }
    \begin{tabular}[t]{c}\bf\rule{\z@}{24\p@}\@author\end{tabular}%
    \vskip 0.3in \@minus 0.1in
  }
}

\renewenvironment{abstract}%
{%
  \vskip 0.075in%
  \centerline{\large\bf Abstract}%
  \vspace{0.5ex}%
  \begin{quote}%
}
{
  \par%
  \end{quote}%
  \vskip 1ex%
}
\makeatother

\usepackage{hyperref}       
\usepackage{url}            
\usepackage{booktabs}       
\usepackage{amsfonts}       
\usepackage{nicefrac}       
\usepackage{microtype}      
\usepackage{xcolor}         
\usepackage{graphicx}
\usepackage{subcaption}
\usepackage{tabularx}
\usepackage{amsmath}
\usepackage{cleveref}
\usepackage{wrapfig}
\usepackage{makecell}

\title{Uncertainty Aware Quality-Diversity Optimization For Offline Biological Sequence Design}

\author{%
  Alona Rimon$^{1}$, \
  Hannah Janmohamed$^{1}$, \
  Maxime Allard$^{2}$, \
  Antoine Cully$^{1}$ 
  \\[1em]
  $^{1}$Imperial College London \\
  $^{2}$Helical
}

\begin{document}

\maketitle 

\begin{abstract} 
Designing sequences such as nucleic acids and proteins with desired properties is a fundamental challenge in biology. Due to the vast, discrete nature of sequence spaces and the high cost of experimental evaluation, offline computational sequence design plays a crucial role in proposing novel sequences before costly wet-lab validation. In practice, it is also essential that proposed designs are diverse to increase the likelihood of downstream experimental success. This motivates methods that balance performance, novelty, and diversity, whilst mitigating against uncertainty inherent to offline optimization. Quality-Diversity optimization (QD) can discover diverse, high-performing solutions, making it well suited to this problem, but its elitist nature makes reliable exploration under uncertainty challenging. To address this, we introduce Prior-guided Offline MAP-Elites (\textbf{POME}), an uncertainty-aware QD framework explicitly adapted to the offline regime. Our method leverages foundation models as priors to guide search toward biologically plausible regions and enable robust exploration under uncertainty. Across three sequence design tasks, POME consistently improves both performance and diversity over prior offline QD methods, and achieves the most reliable balance compared to offline MBO baselines, obtaining the best performance on two tasks and the second-best on the third, while maintaining high diversity and novelty.
\end{abstract}

\section{Introduction}
\label{Introduction}

\begin{figure}[t] 
    \centering
    \begin{subfigure}[t]{0.38\textwidth}
        \centering
        \includegraphics[width=\textwidth]{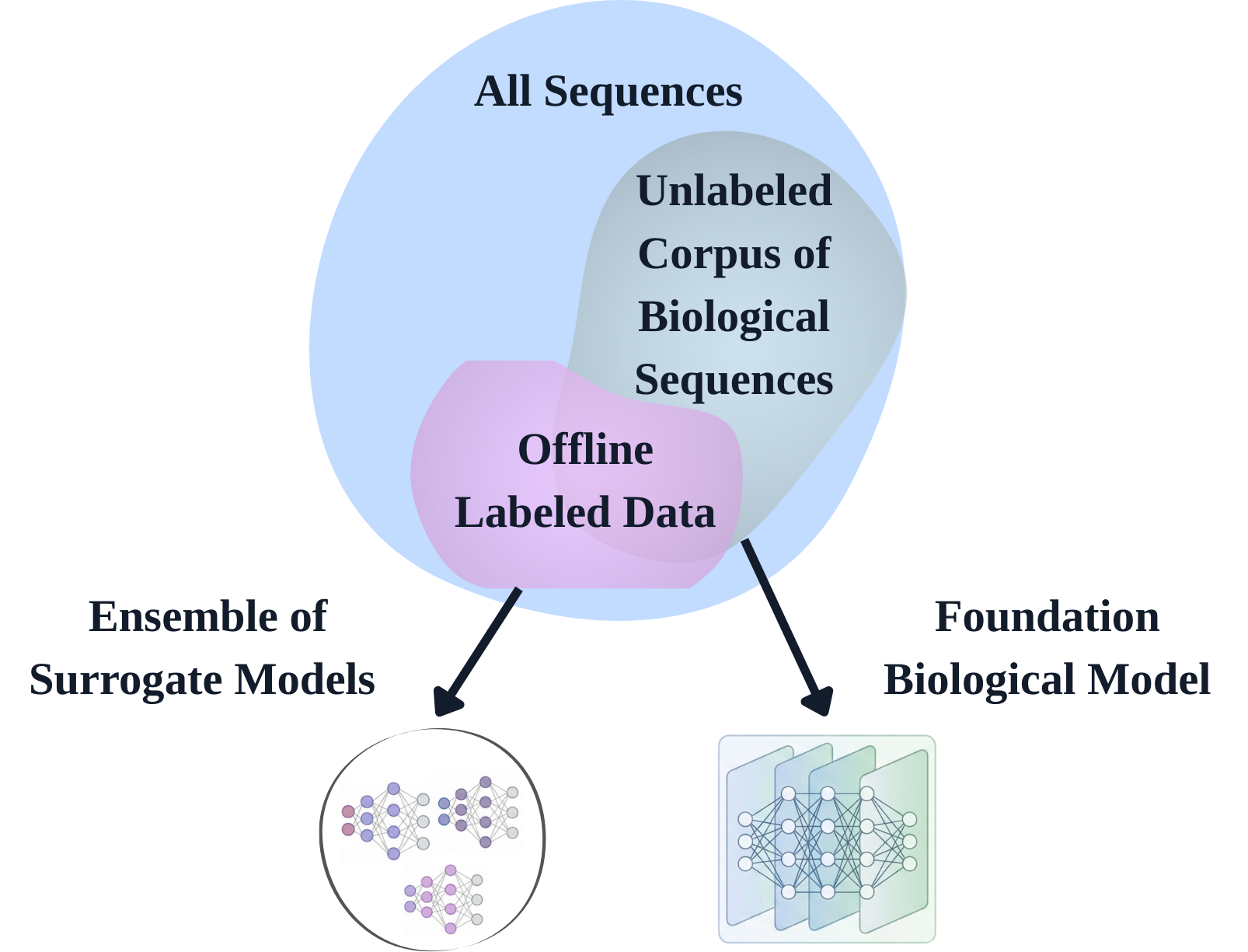}
        \caption{Surrogate and foundation models.}
        \label{fig:offline_vs_fm}
    \end{subfigure}
    \hfill
    \begin{subfigure}[t]{0.56\textwidth}
        \centering
        \includegraphics[width=\textwidth]{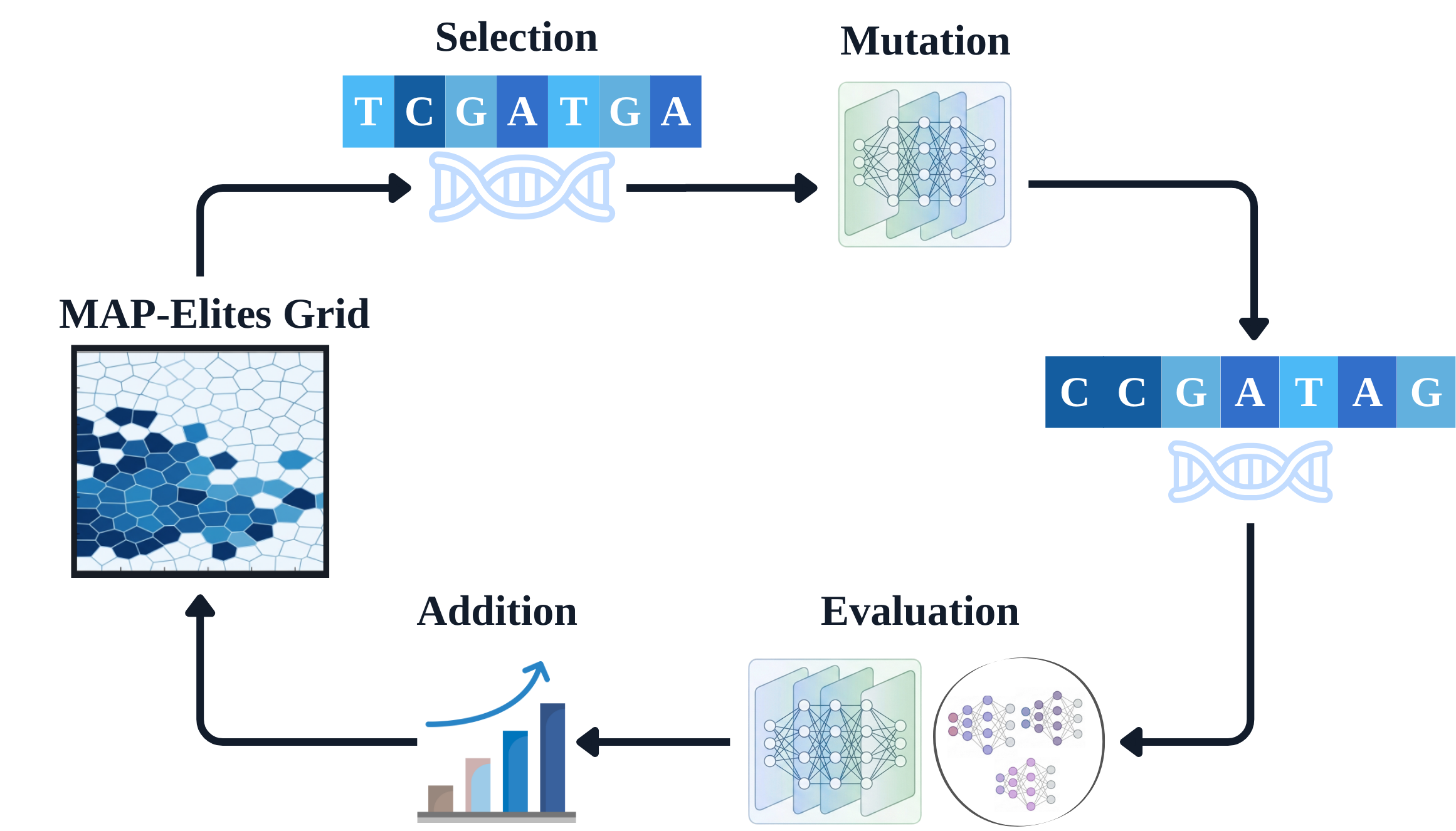}
        \caption{Optimization pipeline.}
        \label{fig:pipeline_loop}
    \end{subfigure}
    \caption{
    Overview of POME, our offline QD optimization framework. 
\textbf{(a)} An ensemble of surrogate models is trained on the offline dataset, 
while a foundation model is pre-trained on large unlabeled biological corpora. 
\textbf{(b)} Sequences are sampled from the archive, mutated using the foundation model, 
evaluated via an uncertainty-aware combination of surrogate predictions and the foundation model prior, 
and reinserted if they improve the corresponding cell elite.
    }
    \label{fig:full_pipeline}
\end{figure}

Designing biological sequences with targeted functional properties, such as high ribosome load or strong binding affinity to transcription factors, is a fundamental challenge in synthetic biology~\cite{salis2009automated_design}, vaccine development~\cite{pardi2018mrna_vaccines} and therapeutic engineering~\cite{fjell2012designing_antimicrobial}. In practice, practitioners often perform multiple rounds of generating variants, building them, testing their properties, and updating the design strategy \cite{nielsen2016engineering_btld}. A major obstacle to this process is that evaluating candidate sequences requires slow and costly wet-lab experiments that can take days to weeks, creating a bottleneck. Computational sequence generation therefore becomes a critical first step, allowing researchers to narrow the search space in silico before committing to experimental validation~\cite{dirks2004paradigms, zhang2023algorithm_for_mrna, cobb2013directed}. This computational generation commonly operates in an offline setting. Only a fixed dataset of designs evaluated in prior high-fidelity in vitro or in vivo experiments is available, while the true target property function remains unknown and expensive to query.  Optimization is typically guided by surrogate models trained on this offline dataset, which approximate the true unknown function and are used to propose new, high-performing designs without access to additional evaluations. Importantly, biological design methods are typically expected to produce sets of diverse candidates rather than a single solution \cite{jeremie24one_shot_qd,jain2022biological_gflownets, kim2025offlineMBO_review}. This need for diversity is essential because many candidates that seem promising computationally may fail in wet-lab validation. Generating a diverse set of sequences that spans multiple functional modes increases the likelihood that at least one will succeed experimentally. Moreover, a diverse set of solutions provides researchers with a richer understanding of the solution space. However, explicitly encouraging diversity introduces an additional optimization challenge, as it requires balancing performance against coverage of multiple plausible modes of the design space. 

Quality-Diversity algorithms naturally address this need for diversity. Rather than seeking a single optimal solution, QD methods aim to generate a diverse set of high-performing candidates, each corresponding to a distinct behavioral mode defined by a user-specified feature vector~\cite{chatzilygeroudis2021qd_bookchapter}. Among QD algorithms, MAP-Elites is widely used for generating diverse, high-performing solution sets by discretising a user-defined feature space and maintaining the best solution per niche \cite{mouret2015map_elites}. However, QD methods typically assume access to an accurate, or at least queryable, fitness function. In most cases, they operate in online settings, where evaluation feedback can be used to iteratively refine the search. By contrast, the setting of offline QD under epistemic uncertainty, where the objective function must be approximated from limited data, remains largely unexplored. A central challenge in this setting is that optimization should ideally discover novel solutions that improve upon those already present in the offline dataset. Achieving such improvements necessarily requires extrapolating beyond the support of the data using surrogate models. However, extrapolation in high-dimensional sequence spaces is inherently uncertain: surrogate models trained on limited data can be highly unreliable for out-of-distribution sequences. Direct optimization against these models often leads to adversarial or biologically invalid designs that exploit model errors rather than reflecting true biological performance \cite{szegedy2013intriguing, trabucco2021coms, kim2025offlineMBO_review}. 

In this work, we address this gap by proposing Prior-guided Offline MAP-Elites (POME), an uncertainty-aware Quality-Diversity approach for offline design. Our method explicitly accounts for epistemic uncertainty in surrogate models when guiding exploration across the feature space. However, uncertainty alone does not specify where to extrapolate safely beyond the offline data: in biological sequence spaces, many out-of-distribution regions are not only uncertain but biologically invalid, making uninformed exploration ineffective. To support controlled extrapolation, we incorporate foundation models as structured priors over the sequence space. These models, pre-trained on large biological sequence corpora, capture salient biological regularities such as codon usage, regulatory structure, and global plausibility~\cite{rives2021biological, consens2025transformers_genome}. 

We show that incorporating foundation models as priors in offline QD optimization helps address the limitations of finite labeled data by injecting prior knowledge over sequence space, guiding search toward more meaningful regions of the search space. We also present POME-MO, a multi objective variant that allows researchers to control the certainty-performance trade-offs of the optimization. We show that explicitly adapting Quality-Diversity optimization to the offline setting is crucial for reliable exploration, and substantially outperforms existing approaches. Compared to prior offline QD methods, our framework consistently improves both performance and diversity across three biological tasks, with gains of $14-37\%$ in performance and $22-35\%$ in diversity, demonstrating that POME enables robust and dependable exploration under epistemic uncertainty. When compared to offline MBO baselines, our method achieves the most consistent balance across tasks, obtaining the best performance on two tasks and the second-best on the third,  while maintaining high diversity and novelty. These results establish our approach as a practical and effective solution for Quality-Diversity optimization in strictly offline regimes. Finally, we show that leveraging foundation models as mutation operators substantially improves search efficiency in high-dimensional discrete design spaces, enabling the discovery of realistic solutions beyond those reachable with uninformed mutations. 

\section{Background}

\paragraph{Offline model-based optimization.}

Offline Model-Based Optimization (MBO) \cite{trabucco2022design_bench, kim2025offlineMBO_review} is applicable across domains where evaluation is costly or constrained, such as materials discovery~\cite{hamidieh2018superconductor}, robotics~\cite{yu2020mopo}, and neural architecture search~\cite{zoph2016neural_with_rl}, as well as biological sequence design~\cite{angermueller2019model_based_rl, brookes2019conditioning, jain2022biological_gflownets}.
In offline MBO, algorithm $A$ aims to output a batch of $K$ candidate solutions, given access only to a static dataset $D = \{(x_i, F_f(x_i))\}_{i=1}^N$ of previously evaluated designs and their corresponding scores. The true target property function $F_f$ is unknown and cannot be queried during optimization. Instead, the algorithm must rely on models learned from the offline data to propose high-performing solutions. Designing bio-sequences in the offline setting presents unique algorithmic and biological challenges. Sequence spaces are combinatorial and discrete. For example, a 50-nucleotide RNA sequence has $4^{50}$ possible combinations, yet only a minute fraction of this space corresponds to viable biological sequences \cite{romero2009exploring, wrenbeck2016plasmid}, making exhaustive search infeasible. At the same time, offline optimization assumes access to a static dataset that is often biased: high-performing sequences tend to lie on a thin, under-sampled manifold \cite{trabucco2022design_bench}.
These difficulties are further compounded by biological fitness landscapes being highly sensitive, rugged, and multimodal \cite{romero2009exploring}, so proximity in sequence space does not imply similar behavior. Without access to new evaluations, surrogate models must extrapolate from limited data, often leading to poor generalization and high epistemic uncertainty. Optimization against such models can exploit prediction errors (objective or reward hacking) widening the sim-to-real gap between computational design and wet-lab validation \cite{trabucco2022design_bench, szegedy2013intriguing, trabucco2021coms}. 

\paragraph{Quality-diversity optimization.} Quality-Diversity optimization is a class of methods that explicitly aim to maintain collections of high-performing solutions that differ in their qualitative behavior~\cite{cully2017qd_unifying_framework, chatzilygeroudis2021qd_bookchapter}.  In many Quality-Diversity algorithms diversity is defined with respect to a user-specified \emph{feature vector}: a mapping that assigns each solution a low-dimensional vector describing salient aspects of its behavior or structure. The choice of feature space determines along which dimensions diversity is encouraged and is typically domain and task dependent. MAP-Elites~\cite{mouret2015map_elites} is a flagship QD algorithm that operationalises this idea. It partitions a user-defined feature space into cells and stores the best-performing solution (the “elite”) in each. For example, in robotics, Cully et al.~\cite{cully2015robots} describe locomotion controllers by the contact patterns of the robot’s legs with the ground. MAP-Elites then discovers diverse walking strategies characterized by different foot-ground contact behaviors, while optimizing locomotion performance within each niche. New candidates are generated by mutating existing elites, evaluated, and inserted if they outperform the current occupant of the corresponding cell, thereby optimizing performance within each niche while encouraging coverage across the feature space. MAP-Elites typically assumes that evaluations of candidate solutions are reliable. In this setting, the elitist update rule is well justified: elites are retained because they reflect true improvements under accurate evaluations. However, this assumption becomes problematic when evaluations are uncertain~\cite{flageat2023uncertain_qd}. In particular, under imperfect evaluations, the elitism mechanism can cause overestimated solutions to be retained as elites, biasing the archive toward regions where surrogate predictions are least trustworthy. Addressing this failure mode is central to adapting Quality-Diversity optimization to the offline settings.

\section{Problem formulation}
\label{Problem Formulation}

We formally define the task of designing diverse biological sequences with specified target properties as an offline model-based quality-diversity optimization problem. Given a finite vocabulary $\mathcal{A}$ (e.g. nucleotides, amino-acids); a sequence space $\mathcal{X} = \mathcal{A}^L$ of all length-$L$ sequences; an unknown true objective function $F_f: \mathcal{X} \rightarrow \mathbb{R}$ (also called the oracle) capturing the desired biological property; an offline labeled dataset $D = \{(x_i, y_i)\}_{i=1}^N$ with $y_i = F_f(x_i)$; a feature function $F_b: \mathcal{X} \rightarrow \mathbb{R}^d$; and a discretized feature space $\mathcal{B}$ partitioned into $K$ cells, the goal is to return a batch of $K$ sequences $\{x_i\}_{i=1}^K$ that achieve high scores under $F_f$ while exhibiting high diversity in the feature space $\{F_b(x_i)\}_{i=1}^K$, ideally covering a significant portion of $\mathcal{B}$. Formally, the optimization target is:
\(
\forall b \in \mathcal{B}, \quad x_b^* = \arg\max_{x \in \mathcal{X}} F_f(x) \quad \text{subject to} \quad F_b(x) \in b.
\)
The returned batch of $K$ sequences aims to include at least one high-performing sequence per feature bin in $\mathcal{B}$. Crucially, optimization is performed in a strictly offline setting, without access to online queries of the oracle function $F_f(x)$.

\section{Method}
\label{Method}

We propose an offline Quality-Diversity algorithm for sequence design, built on the MAP-Elites framework \cite{mouret2015map_elites}. We distinguish between two complementary notions of how atypical a sequence is (illustrated in Figure~\ref{fig:offline_vs_fm}): (i) out-of-distribution with respect to the offline dataset (Section~\ref{uncertainty_estimation}); and (ii) deviation from natural biological sequences, captured independently of the offline dataset  (Section~\ref{prior}). We hypothesize that exploration in natural sequence space should be permitted even when surrogate predictions are moderately uncertain, provided that candidates remain biologically plausible. By incorporating a prior over real-world sequence structure, we bias search away from unrealistic regions while still allowing controlled extrapolation toward high-performing designs. 

\subsection{Algorithmic components and optimization loop}

The proposed algorithm is guided by two core modeling components (see Figure~\ref{fig:offline_vs_fm}):
\textit{(i) a foundation model}, pretrained on large unlabeled corpora of biological sequences, captures global biological regularities and acts as a prior over the sequence space; and
\textit{(ii) an ensemble of surrogate models}, trained on the labeled offline dataset $D$, which provide task-specific performance predictions and epistemic uncertainty estimates. These models are integrated within a MAP-Elites optimization framework that maintains an archive of candidate sequences organized over a discretized feature space using CVT-MAP-Elites \cite{vassiliades2017cvt}. Optimization proceeds as an iterative loop, following the standard QD paradigm \cite{cully2017qd_unifying_framework} (See Figure~\ref{fig:pipeline_loop}). The archive is initialized by inserting a batch of sequences sampled from the offline dataset. At each iteration, the following steps are performed:

\textit{Selection:} $K$ elites are sampled uniformly at random from the current archive.
\textit{Mutation:} Each selected elite is mutated using the foundation model acting as an autoregressive generator. A uniformly random position is selected, and a consecutive subsequence is regenerated, enabling structured, biologically informed exploration in sequence space (\Cref{prior}).
\textit{Evaluation:} Each candidate sequence is assigned a fitness score using a posterior-based heuristic that combines predicted task performance, epistemic uncertainty, and biological plausibility (\Cref{sec:posterior}).
\textit{Addition:} Each candidate is mapped to its corresponding feature cell and replaces the current elite only if it achieves a higher fitness score under the approximated posterior-based evaluation; otherwise, it is discarded.

\paragraph{Prior: sequence likelihood.}\label{prior}

We use a sequence likelihood derived from the foundation model, which remains frozen throughout optimization to preserve its learned inductive biases (see Appendix~\ref{app:foundation_priors}). This model is pre-trained in an autoregressive fashion, using a causal next-token prediction objective~\cite{ferruz2022protgpt2}. The total log-likelihood of a sequence \(x = (x_1, x_2, \ldots, x_T)\) is computed as: $
\log p(x) = \sum_{t=1}^{T} \log p(x_t \mid x_{<t}),$ where \(x_{<t}\) denotes the prefix \((x_1, \ldots, x_{t-1})\). Importantly, this prior is not a Bayesian prior in the traditional sense, but rather a learned proxy for biological plausibility, capturing global constraints learned from real data. It reflects how “natural” the candidate sequence is under the model's learned distribution. While this prior helps guide the search away from biologically implausible regions of sequence space, it provides no information about task-specific performance and is independent of the target property being optimized. We use this prior estimate both (a) to generate new sequences (see Appendix \ref{app:mutate}) and (b) within the fitness function, as described in \cref{sec:posterior}.

\paragraph{Surrogate estimation.}
\label{surrogate_estimation}
\label{uncertainty_estimation}

To estimate task-specific performance and assess the reliability of these predictions for a given candidate sequence, we use an ensemble of independently trained surrogate models, following \cite{jeremie24one_shot_qd}. Concretely, we train an ensemble of $M$ multi-layer perceptrons in a regression setting. For any sequence \(x\), each ensemble member \(f_m(x)\) produces a scalar prediction of task performance. We compute the ensemble mean $\mu(x) = \frac{1}{M} \sum_{m=1}^{M} f_m(x),$ and variance $\sigma^2(x) = \frac{1}{M} \sum_{m=1}^{M} \left( f_m(x) - \mu(x) \right)^2$. The ensemble mean provides an estimate of the expected task performance \(\mathbb{E}[y \mid x]\), while the standard deviation reflects epistemic uncertainty in the surrogate predictions. This uncertainty captures model disagreement arising from limited coverage of the offline dataset in a given region of sequence space~\cite{fort2019deep_ensembles, hastie2009elements}.

\subsection{Fitness estimation: posterior-guided heuristic}
\label{sec:posterior}

In the evaluation stage of the optimization loop (\cref{fig:pipeline_loop}), each candidate sequence is assigned a fitness score using a posterior-guided estimator, which constitutes the core mechanism by which POME handles epistemic uncertainty in the offline setting. Rather than directly maximizing surrogate predictions, fitness evaluation is framed as approximate posterior inference, explicitly balancing predicted performance with biological plausibility under the foundation model prior. Following a Bayesian decomposition and taking logarithms, this score can be expressed, up to an additive constant, as the sum of a prior term and a likelihood term: 
\(
\log p(x \mid y \ge y_{\mathrm{th}}) \propto
\log \underbrace{p(x)}_{\text{prior}} +
\log \underbrace{p(y \ge y_{\mathrm{th}} \mid x)}_{\text{likelihood}} .
\label{eq:posterior}
\) The log-prior \(\log p(x)\) is given by the foundation model as described in~\Cref{prior} and reflects the biological plausibility of a sequence under the model’s learned distribution.  
The likelihood term is defined as the probability that the task performance
$y$ exceeds a threshold $y_{\mathrm{th}}$ given sequence $x$. We set
$
y_{\mathrm{th}} = q\,(y^{\max}_{\text{offline}} - y^{\min}_{\text{offline}})
+ y^{\min}_{\text{offline}},
$
using a percentile-based rule applied consistently across tasks. To instantiate the likelihood term, we assume a Gaussian approximation of the surrogate predictive distribution,
\(
y \mid x \sim \mathcal{N}(\mu(x), \sigma^2(x)),
\)
where \(\mu(x)\) and \(\sigma^2(x)\) are the ensemble mean and variance defined in~\Cref{surrogate_estimation}. Under this Gaussian assumption, the log-probability that the true task performance exceeds the threshold is given by:
\(
\log p\big(y \ge y_{\mathrm{th}} \mid x\big)
= \log\left(1 - \Phi\left(\frac{y_{\mathrm{th}} - \mu(x)}{\sigma(x)}\right)\right),
\label{eq:gauss}
\)
where \(\Phi\) denotes the standard normal cumulative distribution function. 

This evaluation operator is central to POME: by explicitly integrating biological priors, epistemic uncertainty, and task-specific performance into a single scoring function, it enables reliable selection and replacement decisions in offline MAP-Elites. In combination with foundation model guided mutations, this formulation allows POME to prioritize both high-performing and trustworthy regions of the search space, addressing a key failure mode of offline QD methods.

\subsection{POME-MO: Multi-objective QD formulation}
\label{mome_method}

In addition to the approximated posterior-based approach of POME (\Cref{sec:posterior}), we introduce POME-MO, a multi-objective QD formulation. Motivated by recent work in Uncertain QD, we treat performance and predictive certainty as potentially conflicting objectives \cite{flageat2025performence_rep}. 
\begin{wrapfigure}{r}{0.3\linewidth}
    \centering
    \includegraphics[width=\linewidth]{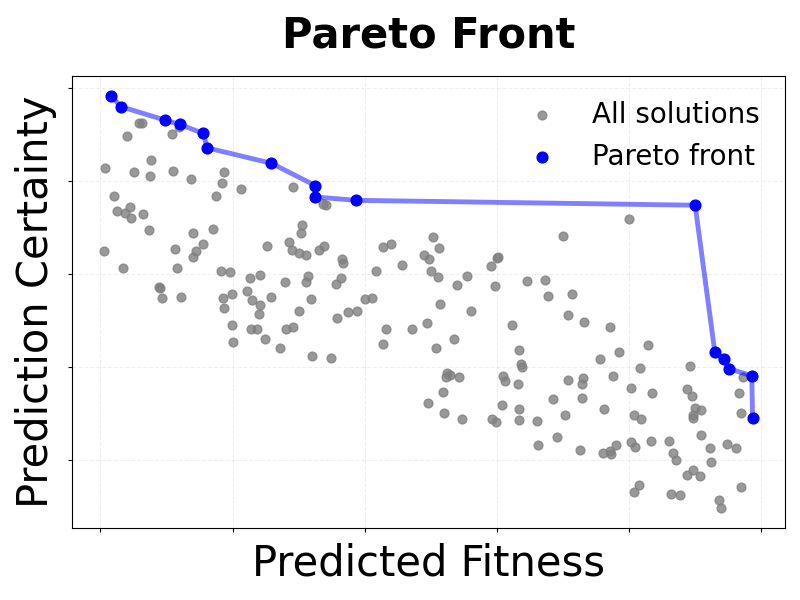}
    \caption{An example of a 2D Pareto front illustrating the trade-off between predicted fitness and certainty across solutions.}
    \label{fig:pareto_front}
\end{wrapfigure}
We adopt a multi-objective MAP-Elites framework that maintains Pareto fronts within each feature cell~\cite{pierrot2022mome}, defined over predicted fitness and certainty (see Section \ref{surrogate_estimation} and \cref{fig:pareto_front}). Importantly, this formulation yields a more interpretable output batch, exposing explicit trade-offs between predicted performance and certainty, rather than collapsing them into a single scalar score. POME-MO shares the same selection and mutation mechanisms as POME, the distinction lies in evaluation and archiving. Instead of assigning a single fitness value, candidates are evaluated along two objectives: predicted fitness ($\mu$) and certainty ($-\sigma^2$). Each feature cell therefore stores a Pareto front rather than a single elite, and new candidates are retained only if they improve this front, with dominated solutions removed. We hypothesize that by preserving solutions that reflect different fitness–uncertainty trade-offs, this formulation allows controlled traversal of high-uncertainty regions when beneficial, facilitating access to otherwise unreachable high-fitness solutions and improving overall coverage relative to fitness-only optimization.

Additional details on POME and POME-MO are provided in Appendices~\ref{app:posterior} and~\ref{app:pome_mo}.

\section{Experimental setup}
\label{sec:exp_setup}
We follow the standardized offline model-based optimization protocol introduced in Design-Bench~\cite{trabucco2022design_bench}. To enable benchmarking without real-world wet-lab experiments, evaluation is performed using an oracle model trained on a hidden full dataset. A restricted offline dataset, obtained by subsampling this full dataset, is provided to the optimization algorithm and constitutes its only source of supervision. The offline MBO method proposes candidate designs using only this offline subset, without access to the oracle or the remaining data. Candidate designs are evaluated post hoc by the oracle solely for benchmarking purposes. An overview of this evaluation protocol is shown in Appendix~\ref{app:exp_setup}, \cref{fig:experimental_framework}. All methods are evaluated using a batch of $K=128$ solutions, excluding sequences from the initial offline dataset. Further details on archive downsampling are provided in Appendix~\ref{app:downsampling}. More information on the experimental setup and tasks can be found in Appendix \ref{app:exp_setup}.

\paragraph{Tasks.}
We evaluate our methods on three biological sequence design tasks from the Design-Bench benchmark suite \cite{trabucco2022design_bench}: 5'UTR, TFBind-10, and GFP. \textbf{The 5'UTR task} involves designing RNA sequences of length 50 to maximize ribosome load. The oracle is a ResNet-based predictor following the Design-Bench protocol. The offline datasets are constructed to emphasise low-performing regions of the search space by retaining only the bottom 50\% of sequences ranked by predicted ribosome load. Mutations and biological plausibility priors are provided by a Helix-mRNA foundation model \cite{wood2025helix}. Feature vectors are defined using nucleotide frequency statistics, as proposed in one-shot QD ~\cite{jeremie24one_shot_qd}. \textbf{The TFBind-10 task} requires designing 10-nucleotide DNA sequences that maximize binding affinity to the \emph{pho4} transcription factor. The dataset originates from the exhaustive binding assay of Barrera et al.~\cite{barrera2016survey}, and the oracle corresponds to an exact lookup table over the full sequence space. The offline dataset consists of sequences sampled uniformly from the bottom $50\%$ of the affinity distribution. We use the Hyena-DNA~\cite{nguyen2023hyenadna} foundation model to guide mutations and impose biological plausibility. Feature vectors are defined based on sequence-space similarity to the offline dataset, following the one-shot QD framework \cite{jeremie24one_shot_qd}. \textbf{The GFP task} focuses on protein sequence design to maximize fluorescence of green fluorescent protein \cite{sarkisyan2016gfp}. We use a ResNet-based oracle and construct the offline dataset by subsampling one third of the full dataset with probability proportional to $\exp(-F_f(x))$, where $F_f(x)$ denotes the oracle-predicted fluorescence score, biasing the dataset toward low-performing sequences. Mutations and prior estimates are guided by the ProtGPT2 protein foundation model \cite{ferruz2022protgpt2}. Feature vectors are defined as in one-shot QD, capturing sequence-space similarity to the offline dataset.

\paragraph{Objectives.}
\label{objectives}
All evaluation metrics are computed with respect to the oracle model, which serves as a proxy for the true task-specific performance. Following the evaluation protocol used in previous works ~\cite{jeremie24one_shot_qd, jain2022biological_gflownets}, for an output batch $B$ of $K$ solutions, we report: \textit{Performance:} Mean and maximum normalized oracle-predicted score over the output batch $B$.
\textit{Diversity:} Average pairwise distance between all proposed solutions in $B$.
\textit{Novelty:} Average distance of each proposed solution to its nearest neighbour in a reference set.

\paragraph{Baselines.}

We compare against a set of standard offline model-based optimization baselines and relevant Quality-Diversity methods. These include \textbf{CbAS} \cite{brookes2019conditioning}, which conditions a generative model on high predicted fitness; \textbf{CMA-ES} \cite{hansen2006cma_es}, an evolutionary strategy that adapts a search distribution using surrogate fitness evaluations; \textbf{COMs} \cite{trabucco2021coms}, a conservative gradient-based offline optimization method that trains conservative surrogates by penalizing overestimation; \textbf{GRAD}, a simple gradient ascent baseline that directly optimizes the surrogate ensemble mean prediction; \textbf{REINFORCE} \cite{williams1992reinforce}, a policy-gradient sequence optimization method; and \textbf{BO-qEI} \cite{wilson2017boqei}, a Bayesian optimization baseline. We additionally include two QD approaches: \textbf{MAP-Elites} \cite{mouret2015map_elites}, using surrogate ensemble predictions to approximate fitness, and \textbf{QD-One-Shot} \cite{jeremie24one_shot_qd}, an uncertainty-aware MAP-Elites variant that penalizes surrogate uncertainty and uses COMs as the surrogate models. For full details of all baselines, see Appendix~\ref{app:baselines}.

\section{Results}

We evaluate all methods on three design tasks, with all reported metrics averaged across 10 independent runs and reported as (mean $\pm$ std).

\subsection{Performance across sequence design tasks}

\textbf{On the UTR task}, POME achieves the strongest overall mean and max performance, while maintaining competitive diversity and novelty (\cref{tab:utr_downsampled_norm}). Offline MBO baselines such as CbAS, COMs, CMA-ES, and REINFORCE attain strong maximum scores and high diversity, but exhibit weaker mean performance, suggesting exploration into unreliable regions of the search space far from the offline data, leading to overestimation.\textbf{ On the TFBind-10 task}, POME achieves the strongest overall performance, attaining the highest maximum score ($0.89$) and the highest mean score ($0.61$), tied with POME-MO (Table~\ref{tab:tfbind10_downsampled_raw_norm}). Among the baselines, REINFORCE and CbAS achieve the highest diversity, but lag behind in mean performance, again suggesting overoptimistic optimization in the offline setting. \textbf{The GFP} landscape is the largest among the evaluated tasks in terms of both sequence length and vocabulary size, and exhibits particularly sharp trade-offs between exploration and reliability in offline optimization (Table~\ref{tab:gfp_downsampled_norm}). REINFORCE and BO-qEI show highly exploitative behavior: given relatively strong initial sequences, they achieve near-optimal scores by collapsing to a narrow region of sequence space, leading to very low diversity and novelty. COMs achieves moderate performance but exhibits unstable behavior, characterized by high variance and very limited diversity. In contrast, CMA-ES explores extremely broadly, achieving by far the highest diversity and novelty, but at the cost of poor performance, suggesting largely unguided exploration. Among methods that achieve more balanced coverage and do not collapse in diversity and novelty, POME and POME-MO attain the strongest overall performance, achieving the highest maximum and mean scores among the remaining approaches. CbAS remains competitive, maintaining high diversity and novelty, but with comparatively lower performance. 

\begin{figure}[t]
    \centering
    \includegraphics[width=0.95\linewidth]{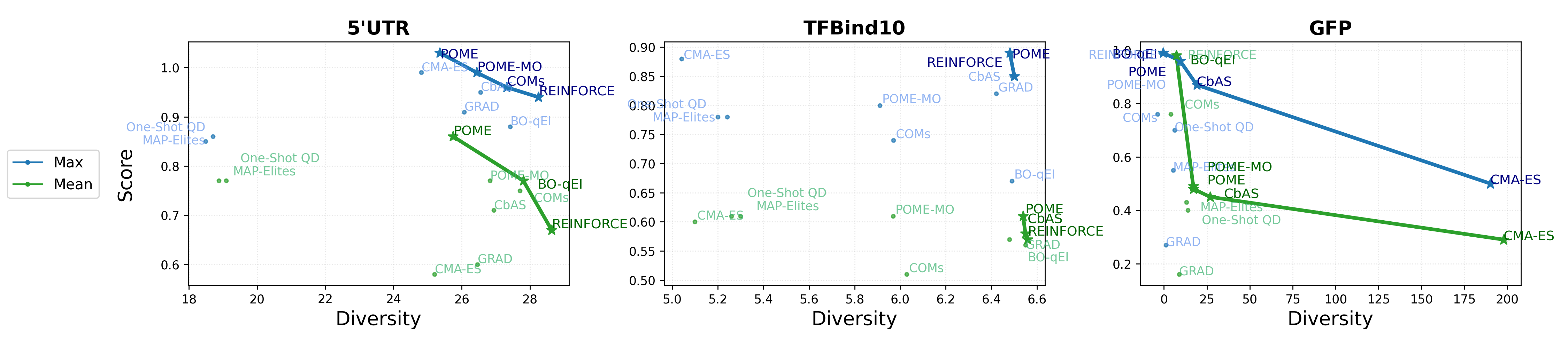}
    \caption{Quality-Diversity trade-offs across tasks (max and mean score vs.\ diversity). Starred points denote Pareto-optimal methods and lines connect the Pareto front.}
    \label{fig:pareto_all_both}
\end{figure}

\begin{figure}[t]
    \centering
    {\includegraphics[width=0.8\linewidth]{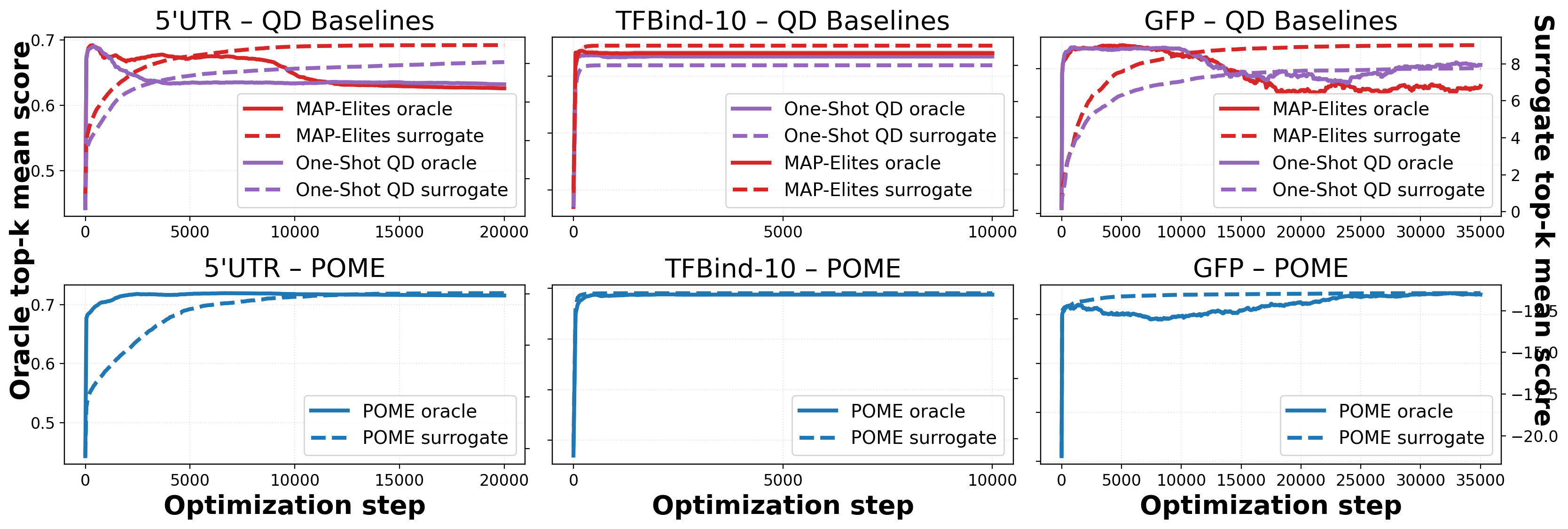}}
    
    \caption{Relationship between surrogate-predicted score and true oracle performance across tasks (mean top-128 scores). Baseline QD methods show misalignment, whereas POME yields closer and more consistent alignment throughout optimization. Note that surrogate scores are on different scales across methods, hence plots are shown separately. Oracle evaluations are performed post hoc for analysis only and are not used at any stage of training, model development, or optimization.}

    \label{fig:oracle_vs_surrogate_all}
\end{figure}


{
{
 \begin{table*}[t]
\centering
\caption{Performance, diversity, and novelty on the 5'UTR task. Results are averaged over 10 runs.}
\small
\label{tab:utr_downsampled_norm}
\begin{tabularx}{\columnwidth}{Xccccl}
\toprule
Name & Max & Mean & Diversity & Novelty \\
\midrule
CbAS & 0.95 $\pm$ 0.01 & 0.71 $\pm$ 0.03 & 26.74 $\pm$ 0.60 & 19.70 $\pm$ 0.13  \\
BO-qEI & 0.88 $\pm$ 0.07 & 0.77 $\pm$ 0.05 & 27.61 $\pm$ 0.69 & 20.64 $\pm$ 0.40  \\
CMA-ES & 0.99 $\pm$ 0.01 & 0.58 $\pm$ 0.04 & 25.01 $\pm$ 1.20 & \textbf{21.24 $\pm$ 0.21}  \\
GRAD  & 0.91 $\pm$ 0.03 & 0.60 $\pm$ 0.01 & 26.26 $\pm$ 0.25 & 19.05 $\pm$ 0.19  \\
REINFORCE & 0.94 $\pm$ 0.02 & 0.67 $\pm$ 0.02 & \textbf{28.44 $\pm$ 0.11} & 19.60 $\pm$ 0.06  \\
COMs & 0.96 $\pm$ 0.02 & 0.75 $\pm$ 0.03 & 27.51 $\pm$ 0.90 & 19.67 $\pm$ 0.20 \\

MAP-Elites & 0.85 $\pm$ 0.01 & 0.77 $\pm$ 0.00 & 18.68 $\pm$ 0.39 & 19.59 $\pm$ 0.06  \\
One-Shot QD & 0.86 $\pm$ 0.02 & 0.77 $\pm$ 0.01 & 18.90 $\pm$ 1.97 & 19.60 $\pm$ 0.08  \\
\midrule
POME & \textbf{1.03 $\pm$ 0.02} & \textbf{0.86 $\pm$ 0.01} & 25.55 $\pm$ 1.38 & 20.24 $\pm$ 0.21   \\
POME-MO & {0.99 $\pm$ 0.00} & 0.77 $\pm$ 0.00 & 26.63 $\pm$ 1.09 & 19.66 $\pm$ 0.15 \\
\bottomrule
\end{tabularx}

\end{table*}

}
{

 \begin{table*}[t]
\centering

\caption{Performance, diversity, and novelty on the TFBind task. Results are averaged over 10 runs.}

\label{tab:tfbind10_downsampled_raw_norm}
\small
\begin{tabularx}{\columnwidth}{Xccccl}
\toprule
Name & Max & Mean & Diversity & Novelty \\
\midrule
CbAS & 0.85 $\pm$ 0.05 & 0.58 $\pm$ 0.01 & 6.52 $\pm$ 0.02 & 2.36 $\pm$ 0.04  \\
BO-qEI  & 0.67 $\pm$ 0.11 & 0.56 $\pm$ 0.00 & 6.52 $\pm$ 0.03 & 2.29 $\pm$ 0.04 \\
CMA-ES & 0.88 $\pm$ 0.04 & 0.60 $\pm$ 0.02 & 5.07 $\pm$ 0.33 & 2.39 $\pm$ 0.08  \\
GRAD & 0.82 $\pm$ 0.06 & 0.57 $\pm$ 0.01 & 6.45 $\pm$ 0.04 & 2.31 $\pm$ 0.06  \\
REINFORCE  & 0.85 $\pm$ 0.03 & 0.57 $\pm$ 0.01 & \textbf{6.53 $\pm$ 0.04} & 2.30 $\pm$ 0.05  \\
COMs & 0.74 $\pm$ 0.02 & 0.51 $\pm$ 0.02 & 6.00 $\pm$ 0.08 & 2.23 $\pm$ 0.04  \\

MAP-Elites & 0.78 $\pm$ 0.02 & 0.61 $\pm$ 0.01 & 5.27 $\pm$ 0.11 & 2.62 $\pm$ 0.05 \\
One-Shot QD &  0.78 $\pm$ 0.02 & 0.61 $\pm$ 0.00 & 5.23 $\pm$ 0.10 & \textbf{2.66 $\pm$ 0.08 } \\
\midrule
POME & \textbf{0.89 $\pm$ 0.02} & \textbf{0.61 $\pm$ 0.01} & 6.51 $\pm$ 0.19 & 2.51 $\pm$ 0.10 \\
POME-MO & 0.80 $\pm$ 0.05 & \textbf{0.61 $\pm$ 0.00} & 5.94 $\pm$ 0.11 & 2.59 $\pm$ 0.07   \\
\bottomrule
\end{tabularx}

\end{table*}
}
{
\begin{table*}[t]
\caption{Performance, diversity, and novelty on the GFP task. Results are averaged over {10} runs.}
\label{tab:gfp_downsampled_norm}
\centering
\small 
\begin{tabularx}{\columnwidth}{Xccccl}
\toprule
Name & Max & Mean & Diversity & Novelty\\
\midrule
CbAS & 0.87 $\pm$ 0.07 & 0.45 $\pm$ 0.02 & 22.99 $\pm$ 1.78 & 10.49 $\pm$ 0.93\\
BO-qEI & \textbf{0.99 $\pm$ 0.01} & \textbf{0.98 $\pm$ 0.00} & 3.32 $\pm$ 0.11 & 1.33 $\pm$ 0.04\\ 

CMA-ES & 0.50 $\pm$ 0.03 & 0.29 $\pm$ 0.01 & \textbf{194.03 $\pm$ 1.3} & \textbf{183.1 $\pm$ 2.12} \\
GRAD & 0.27 $\pm$ 0.14 & 0.16 $\pm$ 0.19 & 4.95 $\pm$ 1.50 & 78.86 $\pm$ 27.58 \\
REINFORCE & \textbf{0.99 $\pm$ 0.00} & \textbf{0.98 $\pm$ 0.00} & 1.98 $\pm$ 0.18 & 0.66 $\pm$ 0.09 \\
COMs & 0.76 $\pm$ 0.30 & 0.76 $\pm$ 0.30 & 0.11 $\pm$ 0.16 & 6.72 $\pm$ 9.51 \\

MAP-Elites & 0.55 $\pm$ 0.10 & 0.43 $\pm$ 0.01 & 9.21 $\pm$ 0.67 & 7.76 $\pm$ 0.17 \\
One-Shot QD &0.70 $\pm$ 0.19 & 0.40 $\pm$ 0.01 & 10.05 $\pm$ 0.96 & 7.39 $\pm$ 0.22  \\ 
\midrule
POME &0.96 $\pm$ 0.02 & 0.48 $\pm$ 0.03 & 12.29 $\pm$ 0.61 & 7.77 $\pm$ 0.25  \\
POME-MO & 0.95 $\pm$ 0.03 & 0.49 $\pm$ 0.02 & 13.61 $\pm$ 0.61 & 7.79 $\pm$ 0.18   \\
\bottomrule
\end{tabularx}

\end{table*}
}
}

To assess offline MBO methods, it is necessary to jointly evaluate solution quality and population-level properties. Figure~\ref{fig:pareto_all_both} plots mean and maximum oracle performance against diversity across all methods and tasks, with starred points denoting Pareto optimal solutions. POME is the only method that Pareto dominates competing approaches in both mean and maximum performance diversity trade-offs across all three datasets. POME attains the highest peak performance on UTR and TFBind and the second-best performance on GFP, while maintaining diverse and stable repertoires. POME-MO further shows that explicitly preserving performance–uncertainty trade-offs within each niche improves coverage in more challenging landscapes.

\subsection{Behavior in comparison to the QD baselines}

POME achieves gains of $14$-$37\%$ in maximum performance and $22$-$35\%$ in diversity over the QD baselines. This suggests that, when applying QD in the offline regime, either without accounting for epistemic uncertainty and biological priors (as in MAP-Elites), or by accounting for uncertainty but without incorporating design-space priors (as in One-Shot QD), optimization becomes biased toward overestimated surrogate predictions. These results are further supported by differences in optimization dynamics between the QD baselines and POME, as shown in Figure~\ref{fig:oracle_vs_surrogate_all}. On the GFP and 5'UTR tasks, MAP-Elites and One-Shot QD exhibit unstable behavior: after an initial improvement over the offline dataset, the true oracle score decreases over optimization, while surrogate predictions continue to increase. This suggests that surrogate overestimation drives the search toward unreliable areas when epistemic uncertainty is not adequately controlled. In contrast, POME maintains both higher and more stable oracle performance throughout optimization, alongside improved alignment between trends in surrogate predictions and true outcomes (we focus on trend alignment rather than absolute prediction values, since MAP-Elites operates via a comparative elite-selection mechanism). This stability translates into stronger archive-level quality and more reliable elite retention. 

\subsection{Extended analysis and ablation studies}
We conduct extensive evaluations to further characterize POME. In Appendix~\ref{app:extended_metrics}, we report additional QD metrics. In Appendix~\ref{app:complexity}, we provide a computational complexity analysis. Furthermore, we conduct ablation studies to isolate the contributions of POME’s key components, including: 
(i) replacing foundation model-guided mutations with random mutations (Appendix~\ref{app:ablation_mutations}), 
(ii) removing the surrogate model and relying solely on the foundation model prior for fitness estimation (Appendix~\ref{app:ablation_surrogate}), 
(iii) removing per-sample uncertainty estimation (Appendix~\ref{app:ablation_uncertainty}),
(iv) evaluating sensitivity to the quantile threshold $q$ (Appendix~\ref{app:ablation_q}), and 
(v) assessing robustness to the choice of uncertainty estimator by replacing the ensemble with a Gaussian Process surrogate (Appendix~\ref{ablation_GP}). We additionally compare two variants, POME-Exp (Appendix~\ref{app:pome_exp}) and POME-Pair (Appendix~\ref{app:pome_pair}), which incorporate uncertainty and priors through alternative fitness formulations.

\section{Related work}

QD methods have shown strong performance in domains such as robotics and materials discovery \cite{cully2015robots, janmohamed2024modq_csp}, but their application to bio-sequence design, particularly in offline settings, remains largely unexplored. To our knowledge, the only work applying QD to offline bio-sequence design is Donà et al.~\cite{jeremie24one_shot_qd}, which adapts MAP-Elites to this regime using Conservative Objective Models \cite{trabucco2021coms} and uncertainty-penalized scoring. More broadly, many QD extensions, such as surrogate-assisted methods including SAIL and BOP-Elites \cite{gaier2017SAIL, kent2023bop}, as well as gradient-informed discrete emitters \cite{boige2023gradient_discrete_qd}, assume repeated oracle access. Most existing QD methods that model uncertainty \cite{flageat2025performence_rep, flageat2023uncertain_qd} are designed for aleatoric uncertainty, where repeated oracle queries allow uncertainty to be reduced through resampling, an assumption that does not hold in offline settings. In parallel, offline model-based optimization has become a prevalent paradigm for biological sequence design, driven by the cost of wet-lab evaluation \cite{trabucco2022design_bench}. These methods rely on surrogate models - ranging from neural networks and ensembles to Conservative Objective Models \cite{trabucco2021coms} and language model–based predictors \cite{chen2023bidirectional}, to approximate the oracle while mitigating OOD overestimation. Candidate sequences are generated through gradient-based optimization \cite{trabucco2021coms}, generative models such as CbAS \cite{brookes2019conditioning}, or model-based reinforcement learning \cite{angermueller2019model_based_rl}. While these approaches achieve strong performance, they typically focus on identifying a single optimum or a small set of high-scoring solutions. In contrast, biological design often requires diverse batches of high-performing candidates, as highlighted in recent works such as GFlowNets~\cite{jain2022biological_gflownets}, DynAMO~\cite{yao2025diversity}, and one-shot QD~\cite{jeremie24one_shot_qd}. This motivates integrating offline model-based optimization with quality-diversity principles. Our work lies at this intersection, combining offline surrogate-based optimization with QD to produce diverse, high-quality sequences under epistemic uncertainty. A more detailed discussion of our positioning relative to multi-objective QD, Bayesian optimization, and online surrogate methods is provided in Appendix \ref{appendix:related_work}.

\section{Conclusion, limitations and future work}
\label{conclusion_limitation_future_work}
We introduced POME, an uncertainty-aware Quality-Diversity framework for offline design that integrates surrogate predictions with epistemic uncertainty and foundation-model priors. Across tasks, POME achieves the strongest balance between performance, diversity, and novelty. Our analysis further shows that uncertainty awareness alone is insufficient in offline QD: without structured priors, QD methods remain vulnerable to surrogate overestimation. By coupling uncertainty-aware selection with generative priors, our framework enables controlled and reliable exploration of high-performing regions of the sequence space under strict offline constraints. Overall, these results indicate that our approach enables robust quality-diversity optimization in offline regimes. Our study has several limitations that point to natural directions for future work. Firstly, our evaluation is benchmark-based with approximated oracles. While standard for offline MBO, bridging the gap to wet-lab validation remains a critical next step for experimental translatability. Specifically, future research should explore multi-stage validation to assess if uncertainty-aware selection improves real-world success rates. Furthermore, the framework could be extended to broader settings, such as handling variable-length sequences and multi-objective optimization. Finally, future work could explore richer feature vectors, account for uncertainty in feature estimation itself, and extend the framework beyond biological sequences to other design domains with structured priors.

\clearpage
\bibliographystyle{plain}
\bibliography{references} 


\clearpage
\appendix
\section{Extended related work}
\label{appendix:related_work}

In this section, we provide a more detailed discussion of the positioning of our work relative to Multi-Objective Quality-Diversity (MO-QD) and online surrogate-assisted QD.

\subsection{Distinctions from state-of-the-art and online QD} \todo{consider shortening this subsection?}
Our work focuses on improving offline optimization under surrogate uncertainty in biological sequence design. Since QD-One-Shot \cite{jeremie24one_shot_qd} is, to our knowledge, the only prior work directly targeting offline QD for biological sequence design, we consider it the most relevant - and we use it as our main baseline alongside MAP-Elites. Most other state-of-the-art methods assume an online or active-learning setting with repeated oracle access. This includes:

\paragraph{Surrogate-assisted quality diversity}
This research line uses surrogate models to improve sample efficiency in expensive domains, typically via an active learning loop: the surrogate guides candidate selection for true oracle evaluation, and the model is then periodically retrained on newly acquired data. For example, SAIL \cite{gaier2017SAIL} utilizes Gaussian Process surrogates and uncertainty-aware acquisition maps to balance exploitation and exploration. Similarly, BOP-Elites \cite{kent2023bop} employs Expected Joint Improvement of Elites (EJIE) to model objectives and descriptors. Crucially, because these methods rely on the iterative refinement of surrogates through repeated oracle queries, they are not directly applicable to the strictly offline regime. \todo{add more relevant citations?}

\paragraph{QD in discrete spaces}
Biological sequence design involves discrete search spaces where standard gradient-based optimization is not directly applicable. ME-GIDE \cite{boige2023gradient_discrete_qd} addresses this by estimating gradients to construct mutation proposal distributions that bias the search toward promising neighbors. However, ME-GIDE assumes an online setup with repeated oracle access. This makes it unsuitable for the strictly offline regime where the model cannot be iteratively updated.

\paragraph{QD under aleatoric uncertainty}
Recent research has explored Quality-Diversity under aleatoric uncertainty, primarily in stochastic environments like robotics or game generation \cite{flageat2023uncertain_qd, flageat2025performence_rep}. While both aleatoric and epistemic uncertainty produce fitness variance, their problem formulations and solution strategies differ fundamentally. Aleatoric methods assume estimates can be refined via repeated oracle queries, an interaction loop that is unavailable in offline settings. Consequently, their evaluation protocols and reliance on resampling are not directly applicable to the strictly offline regime, where uncertainty is epistemic and exploration must be carefully managed to avoid unreliable solutions caused by distribution shift.\todo{add more citations}

\subsection{Multi-objective quality-diversity} \todo{Following the PC's comments: consider making this subsection about multi-objective in general, and then include two parts - in QD and in offline settings (and in both of them to talk about the distinction from our work.}

Several works have explored the intersection of multi-objective optimization and QD. These generally fall into two categories:

\textbf{Diversity as an objective:}  One approach treats novelty and fitness as two objectives but with only local competition between solutions, making it structurally similar to MAP-Elites \cite{lehman2011evolving1}. Other methods utilize specific selection operators \cite{wang2023multi3} or treat quality and diversity as optimization targets, but again within a single-objective fitness framing \cite{do2024evolutionary4}. Crucially, in all three cases the two "objectives" are fitness and diversity, whereas in POME-MO, diversity is handled by the archive structure as in standard QD, and the two objectives are instead predicted performance and surrogate certainty, both of which are fitness-level quantities we seek to jointly optimize. 

\textbf{"True" multi-objective QD:}
Approaches such as multi-objective evolutionary diversity optimization \cite{neumann2019evolutionary2} and multi-objective covariance matrix adaptation \cite{zhao2025multi5} optimize multiple independent fitness objectives within each cell of a QD archive. However, these frameworks are designed for online settings with repeated oracle access. Under strict offline conditions, the benefit of such machinery is ultimately bounded by surrogate reliability rather than the multi-objective strategy itself.

Our approach, POME-MO, differs structurally from these. In POME-MO, diversity is handled by the archive structure as in standard QD, and the two objectives are instead \textit{predicted performance} and \textit{surrogate certainty}. \todo{ADD extended related work ABOUT BO?}

\section{Additional method details}
\label{app:additional_tech_details}

\subsection{Biological foundation models as inductive biases}
\label{app:foundation_priors}
Autoregressive biological foundation models show strong generation and evaluation capabilities. For example, Evo 2 models genomic sequences at scale (\cite{brixi2025Evo2}), while long-context models generate viable bacteriophage genomes validated in wet-lab experiments (\cite{king2025generative}). Such models capture biological structure and can predict functional properties across diverse holdout datasets derived from experimental measurements. This includes HyenaDNA (\cite{nguyen2023hyenadna}), ProtGPT2 (\cite{ferruz2022protgpt2}), Helix-mRNA (\cite{wood2025helix}). These results suggest that autoregressive foundation models provide a powerful inductive bias over biological sequence space, capturing syntactic regularities and functional signals.

\subsection{Posterior-guided heuristic scoring}
\label{app:posterior}

Each candidate sequence $x$ is assigned a scalar fitness score derived from an
approximate posterior over high-performing sequences. The score combines a
task-agnostic biological prior with a task-specific likelihood term derived
from the surrogate ensemble.

\paragraph{Threshold definition.}
We define a performance threshold $y_{\mathrm{th}}$ using a fixed quantile $q$ of
the offline dataset:
\[
y_{\mathrm{th}} = y_{\min}^{\mathrm{off}} + q \bigl( y_{\max}^{\mathrm{off}} - y_{\min}^{\mathrm{off}} \bigr),
\]
where $y_{\min}^{\mathrm{off}}$ and $y_{\max}^{\mathrm{off}}$ denote the minimum and
maximum oracle values in the offline dataset. We use $q=3/4$ in all experiments.

\paragraph{Surrogate likelihood.}
For each sequence $x$, the surrogate ensemble produces a predictive mean $\mu(x)$
and variance $\sigma^2(x)$. We approximate the predictive distribution as
\[
y \mid x \sim \mathcal{N}\bigl(\mu(x), \sigma^2(x)\bigr).
\]
Under this assumption, the likelihood that the true performance exceeds the
threshold is given by
\[
\log p(y \ge y_{\mathrm{th}} \mid x)
= \log \left( 1 - \Phi\left( \frac{y_{\mathrm{th}} - \mu(x)}{\sigma(x)} \right) \right),
\]
where $\Phi$ denotes the standard normal cumulative distribution function.

\paragraph{Final score.}
The final fitness score used for archive insertion is defined as
\[
s(x) = \log p(x) + \log p(y \ge y_{\mathrm{th}} \mid x),
\]
where $\log p(x)$ is the sequence log-likelihood under the frozen foundation model
prior.

While the fitness score $s(x)$ is structured according to a Bayesian decomposition, we clarify that its instantiation in POME serves as a principled posterior-guided heuristic. We distinguish this from strict Bayesian inference in three ways. First, the likelihood is derived from a surrogate ensemble trained on biased offline data. Second, the Gaussian assumption used for the predictive distribution is a heuristic approximation intended to provide a tractable, uncertainty-aware signal for elite replacement. Finally, the foundation model acts as a "biological prior" by capturing structural regularities from vast unlabeled corpora rather than representing a traditional subjective prior belief. This formulation remains a principled way to balance task performance, epistemic uncertainty, and biological validity, which we find is crucial for stable and reliable exploration in strictly offline regimes.

\subsection{Multi-objective QD (POME-MO)}
\label{app:pome_mo}

POME-MO extends POME to a multi-objective quality-diversity setting while preserving
the same feature space, selection strategy, and mutation operators. At each iteration,
elites are sampled uniformly from the archive and mutated using foundation-model-guided
operators, as in the single-objective variant.

\paragraph{Objectives.}
Instead of assigning a single scalar fitness, each candidate sequence $x$ is evaluated
along two objectives: predicted fitness $\mu(x)$ and predictive certainty $-\sigma^2(x)$,
obtained from the surrogate ensemble (\cref{surrogate_estimation}). These objectives
capture a trade-off between performance and reliability under surrogate uncertainty.

\paragraph{Archive structure and insertion.}
Each feature cell maintains a Pareto front over the two objectives rather than a single
elite. A candidate is inserted into a cell if it is non-dominated with respect to the
current Pareto front, in which case any dominated solutions are removed. To preserve diversity along the Pareto front and limit archive growth, we adopt crowding-based replacement mechanisms following Janmohamed et al.~\cite{janmohamed2023improving}, maintaining a maximum Pareto front size of 10 solutions per cell.

\subsection{Downsampling of archives}
\label{app:downsampling}

All methods are evaluated using a fixed output batch size of $K=128$ sequences.
Since MAP-Elites based methods maintain archives whose size depends on the
feature discretisation, a downsampling step is required to produce a final
repertoire of fixed cardinality.

\paragraph{POME.}
For POME we follow the downsampling
procedure introduced in one-shot QD~\cite{jeremie24one_shot_qd}. Specifically,
we construct a Centroidal Voronoi Tessellation (CVT) with $K$ cells using the
empirical distribution of solutions in the final archive, and reinsert all
candidates into this reduced grid, retaining the highest-scoring solution
per cell.

\paragraph{POME-MO: Multi-objective archives.}
This naive downsampling strategy is not suitable for POME-MO, as each cell
contains a Pareto front over predicted fitness and certainty rather than a
single elite. Downsampling solely based on predicted fitness leads to the retention of high-uncertainty solutions and results in an archive dominated by surrogate-driven regions of the search space.

To address this, we first rediscretize the feature space into $K$ CVT cells
using the full set of solutions in the final POME-MO archive. Candidates are
then reinserted into the reduced grid using the posterior-guided fitness score
defined in Section ~\ref{sec:posterior}. This procedure yields a fixed-size
repertoire that reflects a balance between predicted performance, surrogate
uncertainty, and the foundation-model prior.

\subsection{Common implementation details}
\label{app:impl}

All hyperparameters are selected solely based on validation performance on the
offline dataset, in line with the offline model-based optimization setting. No
ground-truth oracle queries beyond the provided offline dataset are used. All
learned surrogate models are trained exclusively on the offline dataset and evaluated on a
held-out validation split using identical optimization settings across methods.

The algorithm was run for 20{,}000 iterations. At each iteration, a batch of 128
genomes was sampled from the repertoire. Each genome in the batch was
independently mutated and reinserted into the archive. The archive is structured
as a CVT grid with 2{,}000 cells, following~\cite{jeremie24one_shot_qd}.

\paragraph{Surrogate ensembles.}
Surrogate predictors are implemented as ensembles of $M = 10$ neural networks.
Following prior work \cite{jeremie24one_shot_qd,jain2022biological_gflownets, trabucco2021coms},
each network consists of two hidden layers of width 2048 with ReLU activations.
We additionally adopt the same relaxed input representation and training
hyperparameters as in these works.

The input to each network is a relaxed, continuous approximation of a one-hot
encoded sequence, given by
\[
\tilde{X} = \log\!\left(CX + \frac{1 - C}{|\mathcal{A}|}\right),
\]
where $C = 0.6$, $X$ denotes the one-hot encoding, and $|\mathcal{A}|$ is the
vocabulary size. Since this transformation introduces a linear dependency
between the coordinates of $\tilde{X}$, the final dimension is removed to
eliminate redundancy.

All surrogate models are trained using the Adam optimizer with a learning rate
of $3 \times 10^{-4}$, batch size 128, and for 50 epochs. Ensemble members are
trained independently with different random initialisations, using a mean
squared error loss.

\paragraph{Mutation operator.} \label{app:mutate}
Offspring are generated by substituting a single position in the parent sequence using a frozen autoregressive foundation model. The mutation location is sampled uniformly at random, and exactly one mutation is applied per offspring. Following one-shot QD~\cite{jeremie24one_shot_qd}, on the GFP task we constrain
mutations to remain within an edit distance of at most 10 from the wild-type sequence. This reflects the structure of the GFP task, which consists almost entirely of sequences mutated by at most 10 positions from the wild type, with the majority lying within a smaller edit distance. If a mutation causes a sequence to exceed this threshold, the mutated position is reverted.

\section{Experimental setup}
\label{app:exp_setup}

\begin{figure}
    \centering
    \includegraphics[width=0.50\linewidth]{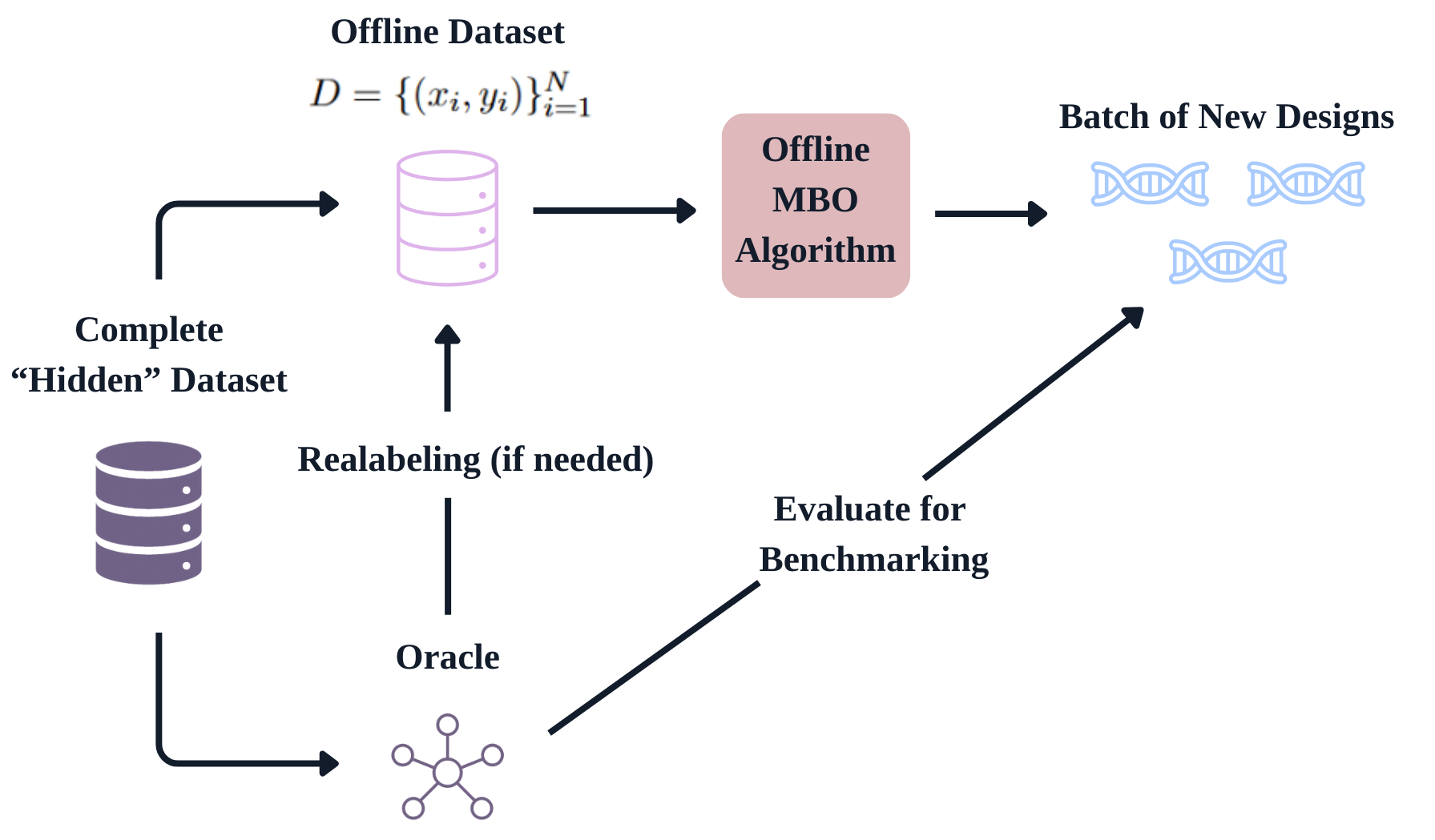}
    \caption{
Overview of the offline model-based optimization evaluation protocol.
A hidden dataset is used to train an oracle, which may be used to relabel the
visible offline dataset provided to the optimization algorithm. The offline MBO
method then generates a batch of candidate designs, which are evaluated by the
oracle for benchmarking.
}
\label{fig:experimental_framework}
\end{figure}

\subsection{Tasks}
\label{app:tasks}

We evaluate our methods on three sequence design tasks drawn from the
Design-Bench benchmark suite~\cite{trabucco2022design_bench}. These tasks
represent distinct biological settings: mRNA translation efficiency, DNA
transcription factor binding, and protein fluorescence. They differ in sequence
length, vocabulary size, offline data subsampling, and oracle structure. Below,
we describe the setup and preprocessing used for each task.

\paragraph{5’ Untranslated Transcribed Region (5'UTR).}
The 5'UTR task involves designing mRNA sequences of length 50 that achieve
maximal ribosome load. The 5' untranslated region (5'UTR) is a non-coding segment
upstream of the coding sequence that strongly influences translation initiation
\cite{neelagandan2020determines}. Ribosome load refers to the number of ribosomes
actively translating an mRNA and is a proxy for protein expression level.
Following the approach of Trabucco et al., we adopt the filtering procedure from
Sample et al.~\cite{sample2019utr}, selecting the top 280{,}000 sequences with the
highest total read counts. This yields a dataset of mRNA sequences of length 50
paired with their corresponding ribosome load values.

\paragraph{Transcription Factor Binding 10 (TFBind-10).}
The TFBind10 task requires designing 10-nucleotide DNA sequences that maximize
binding affinity with the transcription factor \emph{pho4}. Binding affinity
quantifies how strongly a DNA sequence interacts with a transcription factor.
The \emph{pho4} transcription factor is a well-studied yeast regulator. The
dataset originates from the large-scale transcription factor binding activity
survey of Barrera et al.~\cite{barrera2016survey}. Unlike the UTR
task, the full sequence space ($4^{10}$ possibilities) is completely
characterized, and the oracle therefore corresponds to an exact lookup table of
binding affinities. In our preprocessing, we remove duplicate sequences by
averaging their reported affinities.

\paragraph{Green Fluorescent Proteins (GFP).}
The GFP task is based on the Green Fluorescent Protein dataset introduced by
Sarkisyan et al.~\cite{sarkisyan2016gfp}. The dataset studies how mutations of the
\emph{Aequorea victoria} protein affect fluorescence and consists of
approximately 50{,}000 protein variants. All sequences are of fixed length 237
and are obtained through random substitutions relative to the wild-type
sequence. We use the processed version released by Brookes et
al.~\cite{brookes2019conditioning}, following prior
work~\cite{trabucco2021coms,jain2022biological_gflownets,jeremie24one_shot_qd}.

Following one-shot QD~\cite{jeremie24one_shot_qd}, all datasets are min-max
normalized prior to optimization.

\subsection{Oracles}

Following~\cite{trabucco2022design_bench,jeremie24one_shot_qd}, we use an
approximate oracle based on a ResNet architecture for the 5'UTR and GFP tasks,
since the available data does not cover the entire sequence space. The
architecture consists of an embedding layer, a two-layer residual convolutional
network, a self-attention layer, and a final linear projection. Input sequences
are $z$-normalized to zero mean and unit variance. For TFBind10, the oracle is an
exact lookup function, as binding affinities are available for the complete
sequence space.

To verify that our ResNet oracles are comparably reliable to those used
in~\cite{jeremie24one_shot_qd,trabucco2022design_bench}, we evaluate their
performance on held-out validation data. Our oracles achieve Spearman rank correlations
of $0.8707$ for the 5'UTR task and $0.8723$ for the GFP task. These values fall
within $1.5-2.5\%$ of previously reported results. The trained oracles are used to
relabel the sampled offline datasets.

Table~\ref{tab:tasks} summarises the sequence design tasks used in our
experiments, including their sequence space, objectives, oracle models, and
offline datasets.

\begin{table*}[ht]
\caption{Overview of the sequence design tasks used in our experiments.}

\centering
\renewcommand{\arraystretch}{1.3}
\setlength{\tabcolsep}{5pt}
\begin{tabularx}{\textwidth}{|l|
    >{\hsize=0.8\hsize}X|
    >{\hsize=0.8\hsize}X|
    >{\hsize=0.8\hsize}X|
    >{\hsize=1.6\hsize}X|} 
\hline
\textbf{Task} & \textbf{Sequence Space} & \textbf{Objective} & \textbf{Oracle} & \textbf{Offline Dataset} \\
\hline
\makecell[l]{\textbf{5'UTR}} &
50-nt RNA (A, C, G, U) &
Maximize ribosome load &
ResNet-Based &
140,000 samples (bottom 50\%) \newline
Range: \newline
-- 0.1679-0.5924 (data) \newline
-- 0.1632-0.7179 (oracle) \\

\hline
\textbf{TFBind-10} &
10-nt DNA (A, C, G, T) &
Maximize binding affinity (pho4) &
Exact lookup &
100,000 samples (sampled uniformly from bottom 50\%) \newline
Range: \newline
-- 0.0-0.556 (data) \newline
-- 0.0-1.0 (oracle) \\
\hline
\makecell[l]{\textbf{GFP}} &

237-AA Protein
 &
Maximize fluorescence values
 &
ResNet-Based &
18,695 samples (sampled $\propto \exp(-f(x))$) \newline
Range: \newline
--  0.7351-0.9614 (data) \newline
--  0.7351-0.9784 (oracle)
 \\
\hline

\end{tabularx}
\label{tab:tasks}
\end{table*}

\subsection{Baseline details}
\label{app:baselines}

For the classic offline MBO baselines, we use the same architectures and
implementations as defined in the Design-Bench benchmarking work
\cite{trabucco2022design_bench}. Following prior work
\cite{trabucco2022design_bench,jeremie24one_shot_qd,yao2025diversity,jain2022biological_gflownets},
all non-QD baselines (CbAS, BO-qEI, CMA-ES, GRAD, REINFORCE, and COMs) generate multiple  candidate sequences using their respective optimization or sampling procedures. For example, CbAS produces candidates by sampling from a learned generative distribution, whereas GRAD produces the final set of sequences obtained by running multiple gradient-based optimization trajectories in parallel.

For methods utilising MLP surrogate models, the architectures and training protocols (learning rates, number of epochs, and ensemble sizes) are kept identical to those described in Appendix~\ref{app:impl}, ensuring that performance differences arise from the search strategy rather than surrogate capacity.

\paragraph{MAP-Elites.}
We implement a MAP-Elites~\cite{mouret2015map_elites} variant tailored for fair
comparison. We use a random mutation operator in which a position
in the sequence is chosen uniformly at random and a single random substitution is
performed. The surrogate model is an ensemble of ten MLPs trained on the offline
dataset, consistent with the POME variants. Fitness predictions are obtained by
averaging across ensemble members, modeling $\mathbb{E}[y \mid x]$ for a design
$x$ with score $y$. The insertion rule follows the standard MAP-Elites procedure,
replacing the current elite in a cell if the new candidate achieves higher
fitness.

\paragraph{QD-One-Shot.}
We implement the uncertainty-aware MAP-Elites variant for one-shot biological
sequence design proposed by Donà et al.~\cite{jeremie24one_shot_qd}. Unlike
standard MAP-Elites, the surrogate model is an ensemble of ten MLPs trained as Conservative Objective Models (COMs)~\cite{trabucco2021coms}. Each ensemble member produces a point prediction, from which the mean $\mu(x)$ and standard deviation $\sigma(x)$ are computed. The aggregated fitness is penalized by a $\beta$-scaled uncertainty term,
\(
f_{\text{uncertainty-aware}}(x) = \mu(x) - \beta \sigma(x),
\)
encouraging the selection of candidates that achieve high predicted performance while remaining in regions of lower epistemic uncertainty. Random mutation is performed using the same operator as in MAP-Elites.

For fair comparison, both QD-based baselines use the same pipeline parameters as the POME variants, including the number of iterations, number of niches, batch size, feature space definitions, and the maximum mutation constraints applied in the GFP task, as specified in Appendix~\ref{app:impl}.

\section{Extended quality-diversity analysis}
\label{app:extended_metrics}
We provide a deeper analysis of the archive-level performance to further distinguish POME from existing offline QD baselines. Results are averaged over 10 runs (mean $\pm$ std).

The archive profile describes the distribution of performance across the archive by measuring the percentage of cells that contain an elite with an oracle-verified score above a specified threshold $\tau$. Notably, POME shifts a significantly larger portion of the archive into high-fitness regions ($\tau \ge 0.8$) compared to QD baseline approaches.
\label{app:archive}

\begin{table*}[h]
\caption{Archive profile: percentage of cells exceeding oracle thresholds $\tau$.}
\label{tab:archive_profile}
\centering
\begin{tabular}{llcccc}
\toprule
Dataset & Method & $\tau=0.7$ & $\tau=0.8$ & $\tau=0.9$ & $\tau=0.95$ \\
\midrule
\textbf{5'UTR} & MAP-Elites & $\mathbf{97.88 \pm 0.84}$ & $19.06 \pm 1.77$ & $0.01 \pm 0.01$ & $0.01 \pm 0.01$ \\
& One-Shot QD & $95.41 \pm 1.11$ & $19.46 \pm 4.13$ & $0.10 \pm 0.11$ & $0.02 \pm 0.04$ \\
& \textbf{POME} & $95.81 \pm 0.52$ & $\mathbf{85.15 \pm 1.50}$ & $\mathbf{70.36 \pm 5.24}$ & $\mathbf{39.04 \pm 4.92}$ \\
& \textbf{POME-MO} & $89.52 \pm 2.09$ & $36.41 \pm 3.65$ & $5.80 \pm 2.48$ & $1.15 \pm 0.66$ \\
\midrule
\textbf{TFBind} & MAP-Elites & $6.90 \pm 0.46$ & $0.39 \pm 0.07$ & $0.00 \pm 0.01$ & $0.00 \pm 0.00$ \\
& One-Shot QD & $5.99 \pm 0.49$ & $0.35 \pm 0.08$ & $0.01 \pm 0.01$ & $0.00 \pm 0.00$ \\
& \textbf{POME} & $\mathbf{11.96 \pm 1.04}$ & $\mathbf{1.98 \pm 0.26}$ & $\mathbf{0.16 \pm 0.03}$ & $\mathbf{0.01 \pm 0.01}$ \\
& \textbf{POME-MO} & $6.61 \pm 0.33$ & $0.48 \pm 0.06$ & $0.01 \pm 0.01$ & $0.00 \pm 0.00$ \\
\midrule
\textbf{GFP} & MAP-Elites & $5.11 \pm 0.44$ & $4.17 \pm 0.40$ & $2.95 \pm 0.42$ & $1.50 \pm 0.19$ \\
& One-Shot QD & $5.86 \pm 1.19$ & $4.66 \pm 1.06$ & $3.14 \pm 0.86$ & $1.69 \pm 0.54$ \\
& \textbf{POME} & $\mathbf{26.57 \pm 8.38}$ & $\mathbf{19.55 \pm 5.16}$ & $\mathbf{7.61 \pm 1.57}$ & $\mathbf{2.10 \pm 0.69}$ \\
& \textbf{POME-MO} & $18.14 \pm 1.24$ & $13.48 \pm 0.79$ & $5.28 \pm 0.65$ & $1.68 \pm 0.50$ \\
\bottomrule
\end{tabular}
\end{table*}
\section{Computational complexity analysis}
\label{app:complexity}
We provide a detailed breakdown of the computational requirements of POME, distinguishing between one-time and per-iteration costs. Let $K$ be the number of candidates per batch, $M$ the ensemble size, $L$ the sequence length, and $N$ the offline dataset size.

\paragraph{One-time cost:}
\begin{itemize}
    \item Surrogate training: $O(N \cdot M \cdot L)$ (each MLP is 2-layer, constants omitted)
    \item Foundation model: none (pre-trained, frozen).
\end{itemize}

\paragraph{Per iteration:}
\begin{itemize}
    \item Surrogate inference: $O(K \cdot M \cdot L)$.
    \item QD updates: $O(K)$.
    \item Mutation + prior: $O(K \cdot C_{FM}(L))$. For decoder-only transformers, this scales as $C_{FM}(L)=O(N_l(L^2d + Ld^2))$\todo{check complexity}.
\end{itemize}

\paragraph{Comparison to base QD.} Standard MAP-Elites incurs $O(K \cdot M \cdot L)$ for surrogate evaluation, $O(K)$ for QD updates, and $O(K \cdot L)$ for uninformed mutations. Although the foundation model in POME adds compute, it requires no training cost and enables compute-performance trade-offs. Overall, POME's cost scales polynomially in $L$ (linear for surrogates, up to quadratic for transformers), avoiding combinatorial scaling despite the exponential search space. \todo{phrase this better, add about architecture depandency}

While POME introduces additional computational overhead relative to base QD methods, we believe this trade-off is justified in the context of offline biological design. In this regime, the cost of extra compute is relatively low when compared to the high resource and time requirements of wet-lab validation. Since biological evaluation is a multi-stage and expensive process, we consider the allocation of additional compute to improve the reliability of proposed candidates a worthwhile investment. 

Moreover, the QD loop is inherently parallelizable, surrogate and foundation model inference can be batched on GPUs, and frameworks (e.g., QDax) can accelerate execution. We therefore expect substantial speedups with standard parallelization without modifying the algorithm.

\paragraph{Empirical computational cost (UTR):}
Table~\ref{tab:compute_cost} reports the runtime and power consumption averaged over 10 runs (20k iterations) on the UTR task. POME incurs approximately $2\times$ runtime and higher peak power than MAP-Elites, reflecting the foundation model overhead.

\begin{table*}[h]
\centering
\caption{Computational cost on the UTR task (averaged over 10 runs).}
\label{tab:compute_cost}
\begin{tabular}{lcc}
\toprule
Method & Runtime (min) & Peak GPU Power (W) \\
\midrule
MAP-Elites & 20.89 & 192.51 \\
\textbf{POME} & 41.40 & 218.68 \\
\bottomrule
\end{tabular}
\end{table*}

\section{Ablation studies}
\label{app:ablations}

This appendix presents supplementary ablation experiments analysing the
contributions of individual components of POME and POME-MO. Results are reported
separately for each task (5'UTR, TFBind-10, and GFP) as mean and standard
deviation over 5 independent runs, and follow the same evaluation protocol as
the main experiments.

\subsection{Effect of foundation model guided mutations}
\label{app:ablation_mutations}

To assess the role of structured mutations, we replace foundation model guided mutations with uninformed random token substitutions while keeping all other components fixed. Results for both POME and POME-MO across all tasks are reported in Table~\ref{tab:ablation_mutation}.  \todo{add discussion for ablation mutations? (from rebuttal)}

\begin{table*}[h]
\caption{Ablation of foundation-model guided mutations across tasks.}
\label{tab:ablation_mutation}
\centering
\small
\begin{tabular}{llcccc}
\toprule
Task & Method & Max & Mean & Diversity & Novelty \\
\midrule
UTR & POME & {1.03 $\pm$ 0.02} & {0.86 $\pm$ 0.01} & 25.55 $\pm$ 1.38 & 20.24 $\pm$ 0.21  \\
UTR & POME (random mut.) & 1.01 $\pm$ 0.02 & 0.83 $\pm$ 0.02 & 27.45 $\pm$ 1.28 & 20.13 $\pm$ 0.29 \\
UTR & POME-MO & {0.99 $\pm$ 0.00} & 0.77 $\pm$ 0.00 & 26.63 $\pm$ 1.09 & 19.66 $\pm$ 0.15 \\
UTR & POME-MO (random mut.) & 0.96 $\pm$ 0.03 & 0.77 $\pm$ 0.01 & 26.85 $\pm$ 0.83 & 19.68 $\pm$ 0.12 \\
\midrule
TFBind & POME & 0.89 $\pm$ 0.02 & 0.61 $\pm$ 0.01 & 6.51 $\pm$ 0.19 & 2.51 $\pm$ 0.10 \\
TFBind & POME (random mut.) & 0.86 $\pm$ 0.02 & 0.60 $\pm$ 0.01 & 6.63 $\pm$ 0.21 & 2.51 $\pm$ 0.03  \\
TFBind & POME-MO & 0.80 $\pm$ 0.05 & 0.61 $\pm$ 0.00 & 5.94 $\pm$ 0.11 & 2.59 $\pm$ 0.07   \\
TFBind & POME-MO (random mut.) & 0.80 $\pm$ 0.04 & 0.58 $\pm$ 0.01 & 6.28 $\pm$ 0.04 & 2.47 $\pm$ 0.04 \\
\midrule
GFP & POME & 0.96 $\pm$ 0.02 & 0.48 $\pm$ 0.03 & 12.29 $\pm$ 0.61 & 7.77 $\pm$ 0.25  \\
GFP & POME (random mut.) & 0.97 $\pm$ 0.01 & 0.48 $\pm$ 0.03 & 11.66 $\pm$ 0.45 & 7.51 $\pm$ 0.08 \\
GFP & POME-MO & 0.95 $\pm$ 0.03 & 0.49 $\pm$ 0.02 & 13.61 $\pm$ 0.61 & 7.79 $\pm$ 0.18 \\
GFP & POME-MO (random mut.) & 0.96 $\pm$ 0.02 & 0.46 $\pm$ 0.01 & 13.28 $\pm$ 0.24 & 7.58 $\pm$ 0.30 \\
\bottomrule
\end{tabular}
\end{table*}

\subsection{Effect of removing the surrogate model}
\label{app:ablation_surrogate}

We next evaluate variants that remove surrogate-based scoring and rely solely on the foundation model prior. This ablation applies to the posterior-based POME formulation; POME-MO is excluded as it explicitly relies on surrogate-derived objectives. Results across tasks are shown in Table~\ref{tab:ablation_surrogate}.

\begin{table*}[h]
\caption{Effect of removing surrogate-based scoring across tasks.}
\centering
\small
\begin{tabular}{llcccc}
\toprule
Task & Method & Max & Mean & Diversity & Novelty \\
\midrule
UTR & POME & 1.03 $\pm$ 0.02 & 0.86 $\pm$ 0.01 & 25.55 $\pm$ 1.38 & 20.24 $\pm$ 0.21  \\
UTR & Prior-only & 1.00 $\pm$ 0.01 & 0.82 $\pm$ 0.02 & 25.20 $\pm$ 0.35 & 20.62 $\pm$ 0.07\\
\midrule
TFBind & POME & 0.89 $\pm$ 0.02 & 0.61 $\pm$ 0.01 & 6.51 $\pm$ 0.19 & 2.51 $\pm$ 0.10 \\ 
TFBind & Prior-only & 0.86 $\pm$ 0.01 &0.58 $\pm$ 0.01 & 6.74 $\pm$ 0.07 & 2.43 $\pm$ 0.07 \\
\midrule
GFP & POME (full) & 0.96 $\pm$ 0.02 & 0.48 $\pm$ 0.03 & 12.29 $\pm$ 0.61 & 7.77 $\pm$ 0.25  \\
GFP & Prior-only & {0.91 $\pm$ 0.02} & {0.47 $\pm$ 0.02} & {15.27 $\pm$ 0.73} & {7.97 $\pm$ 0.19} \\
\bottomrule
\end{tabular}
\label{tab:ablation_surrogate}
\end{table*}

\subsection{Uncertainty adaptivity}
\label{app:ablation_uncertainty}

We compare adaptive, per-sample uncertainty estimation against a fixed
uncertainty coefficient set to 1, which effectively assumes a normal
distribution with constant variance and does not use the ensemble-predicted
standard deviation on a per-sample basis.  Results across all tasks are reported in
Table~\ref{tab:ablation_uncertainty}.

\begin{table*}[h]
\caption{Effect of adaptive vs.\ fixed uncertainty estimation.}
\centering
\small
\begin{tabular}{llcccc}
\toprule
Task & Method & Max & Mean & Diversity & Novelty \\
\midrule
UTR & POME (full) & 1.03 $\pm$ 0.02 & 0.86 $\pm$ 0.01 & 25.55 $\pm$ 1.38 & 20.24 $\pm$ 0.21  \\
UTR & Fixed uncertainty & {1.03 $\pm$ 0.01} & {0.82 $\pm$ 0.01} & {28.14 $\pm$ 1.00} &{20.46 $\pm$ 0.17} \\
\midrule
TFBind & POME (full) & 0.89 $\pm$ 0.02 & 0.61 $\pm$ 0.01 & 6.51 $\pm$ 0.19 & 2.51 $\pm$ 0.10 \\ 
TFBind & Fixed uncertainty & {0.86 $\pm$ 0.03} & 0.62 $\pm$ 0.00 & 6.20 $\pm$ 0.37 & 2.54 $\pm$ 0.07 \\
\midrule
GFP & POME (full) & 0.96 $\pm$ 0.02 & 0.48 $\pm$ 0.03 & 12.29 $\pm$ 0.61 & 7.77 $\pm$ 0.25  \\
GFP & Fixed uncertainty & {0.79 $\pm$ 0.20} & {0.44 $\pm$ 0.01} & {12.23 $\pm$ 0.45} & {7.80 $\pm$ 0.08}  \\
\bottomrule
\end{tabular}
\label{tab:ablation_uncertainty}
\end{table*}

\paragraph{Discussion.}
First, replacing foundation model guided mutations with uninformed random mutations reduces maximum performance on two of the three tasks (Table~\ref{tab:ablation_mutation}). This indicates that
structured mutations are essential for efficiently navigating high-dimensional
sequence spaces. Second, removing the surrogate model and relying solely on the
foundation model prior for fitness estimation also results in a clear
performance decrease (Table~\ref{tab:ablation_surrogate}). This demonstrates
that biological plausibility alone is insufficient for task-driven optimization,
and that surrogate predictions are critical for directing search toward
high-performing regions. Finally, removing per-sample uncertainty estimation
from the surrogate model (effectively assuming a fixed normal distribution)
leads to a substantial drop in performance, with the maximum GFP score decreasing
from 0.96 to 0.79 and the mean 5'UTR score decreasing from 0.86 to 0.82 (\cref{tab:ablation_uncertainty}). Together,
these results support our central claim that effective offline QD optimization
requires the joint use of structured priors and uncertainty-aware surrogate
guidance.

\subsection{Sensitivity to threshold $y_{th}$ via $q$}
\label{app:ablation_q}
The likelihood estimation in POME depends on the threshold $y_{\text{th}}$ via the quantile parameter $q$. We study its effect by varying $q \in [0.5, 0.8]$ and report results below. 

Overall, performance varies smoothly with $q$, indicating low sensitivity in this range. Key metrics (max and mean fitness) change only marginally (within a few percent), and no single value of $q$ consistently dominates across all metrics or tasks. This suggests that the method is robust to the exact choice of threshold. Importantly, we use a single value of $q$ across all tasks, without task-specific tuning, further supporting robustness to $y_{th}$.

\begin{table*}[h]
\caption{Sensitivity analysis of the quantile parameter $q$ across tasks.}
\centering

\label{tab:q_sensitivity}
\begin{tabular}{lccccc}
\toprule
Task & $q$ & Max & Mean & Diversity & Novelty \\
\midrule
\textbf{5'UTR} & 0.5 & $1.03 \pm 0.01$ & $0.83 \pm 0.00$ & $27.10 \pm 1.60$ & $20.17 \pm 0.38$ \\
& 0.6 & $1.02 \pm 0.01$ & $0.84 \pm 0.03$ & $26.10 \pm 0.55$ & $20.18 \pm 0.10$ \\
& 0.7 & $1.01 \pm 0.00$ & $0.86 \pm 0.01$ & $25.38 \pm 0.67$ & $20.12 \pm 0.07$ \\
& 0.8 & $1.03 \pm 0.01$ & $0.86 \pm 0.02$ & $25.64 \pm 0.97$ & $20.31 \pm 0.14$ \\
\midrule
\textbf{TFBind} & 0.5 & $0.87 \pm 0.03$ & $0.56 \pm 0.01$ & $6.86 \pm 0.06$ & $1.58 \pm 0.01$ \\
& 0.6 & $0.88 \pm 0.02$ & $0.57 \pm 0.02$ & $6.84 \pm 0.07$ & $1.57 \pm 0.05$ \\
& 0.7 & $0.89 \pm 0.02$ & $0.58 \pm 0.01$ & $6.77 \pm 0.11$ & $1.54 \pm 0.06$ \\
& 0.8 & $0.90 \pm 0.02$ & $0.61 \pm 0.01$ & $6.45 \pm 0.19$ & $1.51 \pm 0.05$ \\
\midrule
\textbf{GFP} & 0.5 & $0.94 \pm 0.04$ & $0.46 \pm 0.02$ & $13.04 \pm 0.58$ & $7.39 \pm 1.14$ \\
& 0.6 & $0.92 \pm 0.02$ & $0.47 \pm 0.04$ & $12.81 \pm 0.52$ & $6.68 \pm 0.82$ \\
& 0.7 & $0.96 \pm 0.02$ & $0.51 \pm 0.04$ & $14.85 \pm 0.60$ & $6.37 \pm 0.91$ \\
& 0.8 & $0.95 \pm 0.02$ & $0.45 \pm 0.02$ & $12.74 \pm 0.37$ & $7.16 \pm 0.81$ \\
\bottomrule
\end{tabular}
\end{table*}

\subsection{Robustness to uncertainty estimator: GP Surrogate}
\label{ablation_GP}
We evaluate the impact of replacing the ensemble surrogate with a Gaussian Process (GP). A GP surrogate is trained on the offline dataset: sequences are converted into logits, and both inputs and labels are standardized before fitting. We use a Gaussian Process regressor with a standard RBF kernel, observation noise ($\alpha = 10^{-4}$), and multiple optimizer restarts for stability.

Importantly, as we operate in a strictly offline setting, the GP, like the ensemble, cannot be updated during optimization and is used as a fixed surrogate to provide both mean and variance estimates. We keep all other components of POME unchanged (foundation-model mutations and POME scoring), using the GP as a direct replacement for the ensemble surrogate.

The results in Table \ref{tab:gp_ablation} show that GP-based uncertainty leads to competitive performance across all tasks, with improvements on some tasks (notably GFP) and slight degradation on others (e.g., TFBind mean performance). Results reported as mean $\pm$ std over 5 runs. Overall, these results suggest that POME is not tightly coupled to a specific uncertainty estimator and remains robust across different choices, while benefiting from improved uncertainty estimates. In particular, it can naturally incorporate stronger alternatives such as Gaussian Processes.

\begin{table*}[h]
\caption{Comparison between Ensemble-based and GP-based uncertainty estimation in POME.}
\centering
\label{tab:gp_ablation}
\begin{tabular}{lccccc}
\toprule
Task & Method & Max & Mean & Diversity & Novelty \\
\midrule
\textbf{5'UTR} & POME (Ensemble) & $1.03 \pm 0.02$ & $0.86 \pm 0.01$ & $25.55 \pm 1.38$ & $20.24 \pm 0.21$ \\
& POME (GP) & $1.03 \pm 0.01$ & $0.86 \pm 0.03$ & $23.47 \pm 4.06$ & $20.28 \pm 0.06$ \\
\midrule
\textbf{TFBind-10} & POME (Ensemble) & $0.89 \pm 0.02$ & $0.61 \pm 0.01$ & $6.51 \pm 0.19$ & $2.51 \pm 0.10$ \\
& POME (GP) & $0.91 \pm 0.00$ & $0.56 \pm 0.01$ & $6.26 \pm 0.28$ & $1.62 \pm 0.05$ \\
\midrule
\textbf{GFP} & POME (Ensemble) & $0.96 \pm 0.02$ & $0.48 \pm 0.03$ & $12.29 \pm 0.61$ & $7.77 \pm 0.25$ \\
& POME (GP) & $0.97 \pm 0.00$ & $0.67 \pm 0.06$ & $13.46 \pm 0.58$ & $6.32 \pm 0.35$ \\
\bottomrule
\end{tabular}
\end{table*}

\subsection{Fitness formulations: POME-Pair and POME-Exp}

We further ablate the fitness formulation by examining how the prior over the search space, uncertainty estimation, and task-specific
score prediction, are combined to guide optimization. More specifically, we
introduce and evaluate two additional scoring variants: POME-Pair and POME-Exp.

\subsubsection{POME-Pair: Pairwise ranking (advantage model)}
\label{app:pome_pair}

A key limitation of the Gaussian assumption in the POME method is the need to define an explicit threshold. Interpreting the posterior event  (Section~\ref{sec:posterior}) as "\(y \ge y_{\text{th}}\)", requires introducing a fixed threshold to evaluate a Gaussian tail probability.

However, this absolute thresholding may not be necessary. Since MAP-Elites is inherently an elite-based algorithm, replacement decisions are local: a new solution $x$ only needs to outperform the current elite \(x_{\text{ref}}\) in its respective cell. This suggests a more natural formulation: comparing the candidate directly against the elite, rather than scoring both against a global target.

\paragraph{Why not direct posterior computation.}
In the pairwise setting, an exact posterior formulation would require conditioning on a moving reference sequence whose quality evolves throughout optimization. This introduces a reference-dependent normalisation term that cannot be treated as constant, making direct posterior computation impractical.

\paragraph{Score function formulation.}
Instead of computing an exact posterior, we define a scoring function that combines pairwise evidence with a foundation model prior. At inference time, given a candidate \(x\) and a reference sequence \(x_{\text{ref}}\), we estimate the likelihood of improvement using a learned pairwise model:
\begin{equation}
\log p\big(y(x) \ge y(x_{\mathrm{ref}}) \mid x, x_{\text{ref}}\big),
\label{eq:gain}
\end{equation}
where this likelihood is estimated by a learned pairwise ranking model.

\paragraph{Adaptive uncertainty-aware scoring.}
\label{adaptive_gamma}
To combine task-specific scoring with biological plausibility in a reliability-aware manner, we combine the pairwise likelihood with the foundation model prior using an uncertainty-dependent weight:
\setlength{\abovedisplayskip}{6pt}
\setlength{\belowdisplayskip}{6pt}
\begin{equation}
\begin{aligned}
f_{\mathrm{gain}}(x, x_{\mathrm{ref}}) &=
\bigl(1 - \gamma(x, x_{\mathrm{ref}})\bigr)\,
\log p\!\bigl(y(x) \ge y(x_{\mathrm{ref}})
\mid x, x_{\mathrm{ref}}\bigr) \\
&\quad + \gamma(x, x_{\mathrm{ref}})\,\log p(x).
\end{aligned}
\label{eq:f_gain_adaptive}
\end{equation}

where \(\gamma(x, x_{\mathrm{ref}}) \in [0, 0.5]\) is a sample-specific weighting factor that increases with predictive uncertainty.

\textbf{Adaptive \(\gamma(x)\) key idea:}
\begin{itemize}
  \item When the surrogate is confident (low uncertainty), we trust its output: \(\gamma(x) \rightarrow 0\)
  \item When the surrogate is uncertain, we fall back to the prior: \(\gamma(x) \rightarrow 0.5\)
\end{itemize}

This adaptive weighting ensures that when the surrogate is confident about the pair \((x, x_{\mathrm{ref}})\) (i.e., \(\sigma(x, x_{\mathrm{ref}})\) is low), the score relies more on the pairwise likelihood. Conversely, under high uncertainty, the influence of the prior is increased to avoid over-reliance on potentially unreliable predictions.

We estimate uncertainty using the standard deviation \(\sigma(x)\) across the ensemble predictions. To ensure comparability across tasks with different scales of \(y\), we normalise this uncertainty using min–max statistics:

\begin{equation}
\begin{aligned}
\gamma(x, x_{\mathrm{ref}})
&= \operatorname{clip}_{[0,0.5]}\!\left(
\frac{\sigma(x, x_{\mathrm{ref}}) - \min\limits_{(x', x'_{\mathrm{ref}})\in \mathcal{D}_{\text{offline}}} \sigma(x', x'_{\mathrm{ref}})}
{\max\limits_{(x'', x''_{\mathrm{ref}})\in \mathcal{D}_{\text{random}}} \sigma(x'', x''_{\mathrm{ref}})}
\right. \\
&\qquad \left.
- \min\limits_{(x', x'_{\mathrm{ref}})\in \mathcal{D}_{\text{offline}}} \sigma(x', x'_{\mathrm{ref}})
\right)
\end{aligned}
\end{equation}

where the minimum is computed over the offline dataset and the maximum over a batch of randomly generated sequences.

We cap \(\gamma(x)\) at 0.5 to avoid overly relying on the prior and retain task-specific signal. We also clip \(\gamma(x)\) from below at 0 to avoid invalid or negative weighting when the predicted uncertainty falls below the estimated offline minimum. This maintains numerical stability and preserves the interpretation of \(\gamma(x)\) as a soft trust coefficient between the surrogate and the prior.

\paragraph{Symmetric comparison for elite replacement.}
At selection time, we evaluate both directions to ensure a fair comparison:
\setlength{\abovedisplayskip}{6pt}
\setlength{\belowdisplayskip}{6pt}
\(s_{\text{candidate}} = f_{\mathrm{gain}}(x, e)\),  
 \(s_{\text{elite}} = f_{\mathrm{gain}}(e, x)\)
We replace the elite \(e\) with candidate \(x\) if:
\(
s_{\text{candidate}} > s_{\text{elite}}
\)

This bidirectional evaluation accounts for both the evidence that \(x\) outperforms \(e\), weighted by the plausibility of \(x\), and the evidence that \(e\) outperforms \(x\), weighted by the plausibility of \(e\).

\subsubsection{POME-Exp: expectation-guided optimization}
\label{app:pome_exp}

In this variant, the fitness of a candidate sequence is computed as a weighted combination of its predicted performance and its biological plausibility. Specifically, we combine the expected fitness \(\mathbb{E}[y \mid x]\) (estimated by the surrogate) with the log-likelihood under a foundation model prior:
\setlength{\abovedisplayskip}{6pt}
\setlength{\belowdisplayskip}{6pt}
\[
f(x) = (1 - \gamma(x)) \cdot \mathbb{E}[y \mid x] + \gamma(x) \cdot \log p(x),
\]
where \(\gamma(x)\) is an adaptive weighting factor that adjusts the relative influence of the prior and the surrogate depending on the model's confidence in its prediction.

We use the same adaptive scheme for computing \(\gamma(x)\) as described previously in Section~\ref{adaptive_gamma}. In brief, \(\gamma(x)\) is derived from the normalized uncertainty of the surrogate ensemble: high uncertainty leads to a higher \(\gamma(x)\), increasing reliance on the prior. This allows the optimization to fall back on biologically plausible regions when the surrogate is unreliable, and focus on performance when the surrogate is confident.

\paragraph{Scale normalisation.}
The two components \(\mathbb{E}[y \mid x]\) and \(\log p(x)\) are not naturally on the same scale, so we apply separate normalisation procedures before combination.

The surrogate model is trained on a dataset normalised to the range \([0,1]\), so its predictions typically fall within or near this interval. However, since the predictor is not explicitly constrained to the \([0,1]\) range, the output \(\mathbb{E}[y \mid x]\) is not strictly bounded. In practice, most predictions fall below 1, as the offline dataset primarily contains suboptimal designs. We therefore clip \(\mathbb{E}[y \mid x]\) to the range \([0,1]\).

The log-prior \(\log p(x)\) can vary significantly and is normalised using a min–max procedure analogous to that used for adaptive \(\gamma(x)\):
\[
\log p_{\text{norm}}(x) = \frac{\log p(x) - \min_{x' \in \mathcal{D}_{\text{random}}} \log p(x')}{\max_{x'' \in \mathcal{D}_{\text{offline}}} \log p(x'') - \min_{x' \in \mathcal{D}_{\text{random}}} \log p(x')}.
\]
The result is clipped to the \([0,1]\) range.

This normalisation ensures that the prior and expected performance are weighted on a consistent scale, allowing the adaptive weighting via \(\gamma(x)\) to be meaningful and stable across candidates.

\subsubsection{Results}
\begin{table*}[h]
\caption{Effect of alternative fitness formulations (POME-Pair and POME-Exp)
across all tasks. Results are reported as mean $\pm$ standard deviation over 5
runs.}
\centering
\small
\begin{tabular}{llcccc}
\toprule
Task & Method & Max & Mean & Diversity & Novelty \\
\midrule
UTR & POME-Pair & 0.85 $\pm$ 0.03 & 0.75 $\pm$ 0.01 & 22.65 $\pm$ 1.56 & 19.74 $\pm$ 0.10 \\
UTR & POME-Exp  & 0.90 $\pm$ 0.04 & 0.78 $\pm$ 0.02 & 20.94 $\pm$ 1.25 & 19.36 $\pm$ 0.18 \\
\midrule
TFBind & POME-Pair & 0.78 $\pm$ 0.03 & 0.62 $\pm$ 0.00 & 5.14 $\pm$ 0.19 & 2.67 $\pm$ 0.06 \\
TFBind & POME-Exp  & 0.83 $\pm$ 0.03 & 0.58 $\pm$ 0.01 & 6.16 $\pm$ 0.18 & 2.53 $\pm$ 0.05 \\
\midrule
GFP & POME-Pair & 0.90 $\pm$ 0.03 & 0.43 $\pm$ 0.03 & 9.93 $\pm$ 0.47 & 7.96 $\pm$ 0.19 \\
GFP & POME-Exp  & 0.67 $\pm$ 0.21 & 0.43 $\pm$ 0.01 & 10.12 $\pm$ 0.69 & 7.62 $\pm$ 0.29\\
\bottomrule
\end{tabular}

\label{tab:ablation_fitness_formulations}
\end{table*}

Table~\ref{tab:ablation_fitness_formulations} reports results for the proposed
fitness formulation variants, POME-Pair and POME-Exp, across all tasks. Both
variants underperform the full POME formulation overall, particularly in terms
of stability across tasks. POME-Pair achieves strong performance on the GFP task
but performs less well on TFBind-10, while POME-Exp exhibits the opposite trend,
performing comparatively better on TFBind-10 but worse on GFP. 

Despite this reduced stability, both variants remain highly competitive when
compared to QD baselines such as MAP-Elites and one-shot QD. These results
suggest that incorporating biological priors and uncertainty-aware surrogate
guidance, even through simplified or alternative fitness formulations, already
provides gains over prior offline QD approaches.
\section{Broader impacts}
\label{app:broader_impacts}
This paper presents work whose goal is to advance the field of Machine Learning, specifically in offline quality-diversity optimizations. While the methods studied here are evaluated on benign benchmark tasks
and rely entirely on offline data and computational oracles, sequence design
methods could in principle be misused if applied to harmful objectives.
However, deploying such applications would require substantial domain-specific
expertise and experimental infrastructure (e.g.\ wet-lab facilities), which is beyond the scope of this work. We therefore focus exclusively on controlled,
non-harmful settings and established benchmarks, and do not foresee immediate
negative societal consequences arising from the results presented here.


\end{document}